\documentclass[sigconf]{aamas}  % do not change this line!

\usepackage{booktabs}
\usepackage{algorithm}
\usepackage{algorithmic}
\usepackage{amsmath}

\DeclareMathOperator*{\argmax}{arg\,max}

\usepackage{flushend}
\setcopyright{ifaamas}  % do not change this line!
\acmDOI{doi}  % do not change this line!
\acmISBN{}  % do not change this line!
\acmConference[AAMAS'20]{Proc.\@ of the 19th International Conference on Autonomous Agents and Multiagent Systems (AAMAS 2020), B.~An, N.~Yorke-Smith, A.~El~Fallah~Seghrouchni, G.~Sukthankar (eds.)}{May 2020}{Auckland, New Zealand}  % do not change this line!
\acmYear{2020}  % do not change this line!
\copyrightyear{2020}  % do not change this line!
\acmPrice{}  % do not change this line!

\begin{document}

\title{FRESH: Interactive Reward Shaping in High-Dimensional \\State Spaces using Human Feedback}  
%\title{FRESH: Feedback Neural Network-based Reward Shaping in \\High-Dimensional State Spaces}% put your title here!
%\title{INCREASE: Deep Interactive Reward Shaping in \\High-Dimensional State Spaces using Human Feedback}
%\title{HANDOUT: Deep Interactive Reward Shaping in \\High-Dimensional State Spaces using Human Feedback}
%\titlenote{Produces the permission block, and copyright information}

% AAMAS: as appropriate, uncomment one subtitle line; check the CFP
%\subtitle{Extended Abstract}
%\subtitle{Blue Sky Ideas Track}
%\subtitle{JAAMAS Track}
%\subtitle{Demonstration}
%\subtitle{Doctoral Consortium}

\author{Baicen Xiao}
\affiliation{%
  \institution{University of Washington}
  %\streetaddress{P.O. Box 1212}
  \city{Seattle, WA, USA} 
%  \state{WA} 
%  \postcode{98195}
}\email{bcxiao@uw.edu}
\author{Qifan Lu}
\affiliation{%
  \institution{University of Washington}
  %\streetaddress{P.O. Box 1212}
  \city{Seattle, WA, USA} 
%  \state{WA} 
%  \postcode{98195}
}\email{lqf96@uw.edu}
\author{Bhaskar Ramasubramanian}
\affiliation{%
  \institution{University of Washington}
  %\streetaddress{P.O. Box 1212}
  \city{Seattle, WA, USA} 
%  \state{WA} 
%  \postcode{98195}
}\email{bhaskarr@uw.edu}
\author{Andrew Clark}
\affiliation{%
  \institution{Worcester Polytechnic Institute}
  %\streetaddress{P.O. Box 1212}
  \city{Worcester, MA, USA} 
%  \state{MA} 
%  \postcode{98195}
}\email{aclark@wpi.edu}
\author{Linda Bushnell}
\affiliation{%
  \institution{University of Washington}
  %\streetaddress{P.O. Box 1212}
  \city{Seattle, WA, USA} 
%  \state{WA} 
%  \postcode{98195}
}\email{lb2@uw.edu}
\author{Radha Poovendran}
\affiliation{%
  \institution{University of Washington}
  %\streetaddress{P.O. Box 1212}
  \city{Seattle, WA, USA} 
%  \state{WA} 
%  \postcode{98195}
}\email{rp3@uw.edu}

\renewcommand{\shortauthors}{B. Xiao et al.}

\begin{abstract}  % put your abstract here!
Reinforcement learning has been successful in training autonomous agents to accomplish goals in complex environments. 
Although this has been adapted to multiple settings, including robotics and computer games, human players often find it easier to obtain higher rewards in some environments than reinforcement learning algorithms. 
This is especially true of high-dimensional state spaces where the reward obtained by the agent is sparse or extremely delayed. 
In this paper, we seek to effectively integrate feedback signals supplied by a human operator with deep reinforcement learning algorithms in high-dimensional state spaces. 
We call this \textbf{\emph{FRESH (Feedback-based REward SHaping)}}. 
%We use human feedback to interactively shape the reward received by a reinforcement learning agent in high-dimensional spaces. 
During training, a human operator is presented with trajectories from a replay buffer and then provides feedback on states and actions in the trajectory. 
In order to generalize feedback signals provided by the human operator to previously unseen states and actions at test-time, we use a \emph{feedback neural network}. 
We use an ensemble of neural networks with a shared network architecture to represent model uncertainty and the confidence of the neural network in its output. 
The output of the feedback neural network is converted to a shaping reward that is augmented to the reward provided by the environment. 
We evaluate our approach on the \emph{Bowling} and \emph{Skiing} Atari games in the arcade learning environment. 
Although human experts have been able to achieve high scores in these environments, state-of-the-art deep learning algorithms perform poorly.  
We observe that \emph{FRESH} is able to achieve much higher scores than state-of-the-art deep learning algorithms in both environments. \emph{FRESH} also achieves a $\mathbf{21.4\%}$ higher score than a human expert in \emph{Bowling} and does as well as a human expert in \emph{Skiing}. 
\end{abstract}

%%HANDOUT - HumAN feeDback cOnfidence Using neTworks
  % put your semicolon-separated keywords here!

\maketitle

%%%%%%%%%%%%%%%%%%%%%%%%%%%%%%%%%%%%%%%%%%%%%%%%%%%%%%%%%%%%%%%%%%%%%%%%%%%%%%%%%%%%%%%%%%%%%%%%%%%%%%%%%
%% start of main body of paper

\section{Introduction} \label{Sec:Introduction}

Reinforcement learning (RL) is a framework that enables an agent to explore an environment in order to maximize its expected long-term reward, where the reward signal is supplied by the environment. 
A major attraction of this approach is that the agent can learn to complete tasks %and achieve goals 
even when a model of the environment (i.e. transition probabilities between states) is unknown. 
With adequate computational resources, model-free RL algorithms have been successfully applied to many domains, including board and computer games \cite{mnih2015human, silver2016mastering} and robotics \cite{hafner2011reinforcement, lillicrap2016continuous, schulman2015trust}. 
However, it has been noted that the sample complexity of these algorithms is typically very high, which limits their use on real-world systems \cite{hafner2011reinforcement, mnih2015human}. 
Model-based RL, on the other hand, aims to learn a model of the system, and then optimize a policy given this model \cite{berkenkamp2017safe, gu2016continuous, kamalapurkar2016model, levine2016end}. 
This approach has the benefit of reducing sample complexity, but the scope of the policy is limited by assumptions made on the model. 
%A model-based policy search method that overcame the effects of model bias was presented in \cite{deisenroth2011pilco}. 

Although RL algorithms have shown impressive results in certain environments \cite{mnih2015human, silver2016mastering}, humans are usually much more efficient in terms of the number of actions required to obtain higher rewards. %needed to achieve success. 
This has especially been observed in environments with high-dimensional state spaces, like video games, where states are raw pixels of images or snapshots of videos. 
In these settings, any prior knowledge that a human might have about the environment, and their ability to learn from the environment, is crucial in determining success. 
The authors of \cite{pathak2017curiosity} showed that a human player is easily able to play and win games in setups where the reward structure is sparse or significantly delayed (for e.g. receiving a reward only for successfully completing one level of a game), while deep RL algorithms struggle. 
The role of a human operator in providing a \emph{shaping} reward signal to aid the learning process of the RL agent in such environments has not been addressed in prior work.
%This is more evident in setups where the reward structure is sparse or significantly delayed (for e.g. receiving a reward only for successfully completing one level of a game). 
%A human player is easily able to play and win games in such a scenario, while an autonomous agent struggles \cite{pathak2017curiosity}. 

\begin{figure}
	\includegraphics[width=2.8 in]{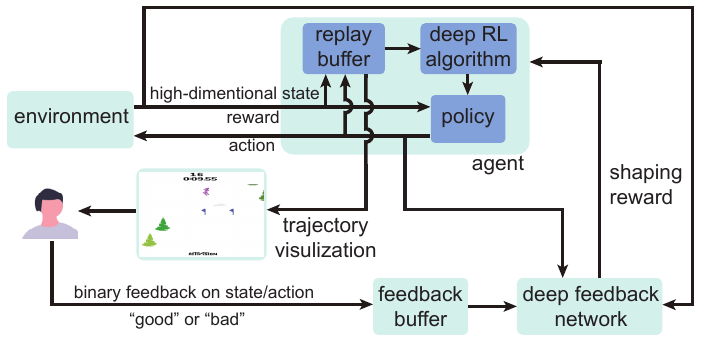}
	\caption{Schematic for \emph{FRESH (Feedback-based REward SHaping)}. A human operator is presented with trajectories of game-play from the replay buffer. At each state in the trajectory, the operator indicates whether the action taken in that state is \emph{good} or \emph{bad}. At some states, the operator also indicates whether the state is \emph{good} or \emph{bad}. This feedback is stored in a feedback buffer. A deep feedback neural network is used to allow the deep reinforcement learning algorithm to generalize feedback signals obtained during training to unseen states and actions at test-time. The output of the feedback network is converted to a shaping reward, which is augmented to the reward provided by the environment.}\label{schematic}
\end{figure}

In this paper, we seek to effectively integrate human feedback with deep RL algorithms in high-dimensional state spaces. 
We term this \textbf{\emph{FRESH}}, for \textbf{Feedback-based REward SHaping}. 
We assume that there is a human operator who can provide feedback on actions taken by the RL agent. 
During training, the operator is presented with trajectories (sequences of states and actions) from a replay buffer and indicates whether an action at a state in the trajectory is \emph{good} or \emph{bad}. 
The operator is additionally able to provide this feedback on the states themselves. 
They can also indicate that they are not sure if an action is good or bad (\emph{cannot tell}). If they feel that sufficient number of feedback signals have been given on a trajectory, they can choose to \emph{skip} the remainder of that trajectory. 
The motivation for this type of feedback is that it is relatively easier for a human to understand whether a state and the consequence of an action taken at a state will be good or bad. 
A schematic of our setup is shown in Figure \ref{schematic}.

We wish to clarify that this approach is different from that used in Deep-TAMER \cite{warnell2018deep}, which used human-feedback, but did not use the environment reward, and instead adopted a supervised learning framework during training. There, a human operator associated a numerical value indicative of how good the agent's behavior was, in the opinion of the operator. 
In comparison, in \emph{FRESH}, the operator only needs to provide a qualitative indication of whether an action in a state is good or bad, and we convert this to a shaping reward that is integrated with the reward from the environment. 
%Deep-TAMER, on the other hand does not use the environment reward, and instead adopts a supervised learning framework during training.

Due to the size of the state space when operating in high dimensional state spaces, 
%A challenge while operating in high-dimensional state spaces is that due to the size of the state space, 
it will not be uncommon for the agent to encounter states at test time that it would have never seen during training. 
Moreover, the number of feedback signals that can be provided by a human is limited. 
Therefore, we need to be able to generalize from the feedback in order to determine whether a state encountered at test time is `similar' to some state seen previously. 
Neural networks (NNs) will allow us to generalize from feedback. 
%A neural network (NN) yields an output for any input given to it. 
Since an NN yields an output for any input given to it, we require a mechanism to reject outputs that could lead the RL agent towards `bad' states. 
One way to achieve this is by quantifying a measure of confidence of the network in its output. 
This way, by setting an appropriate threshold, outputs of the network that have a confidence score below this threshold can be rejected. Only those outputs with a confidence score above the threshold will be retained and used during the training phase. 
%This way, by setting an appropriate threshold, a fraction of the outputs of the network can be rejected, and only those outputs retained will be used during the training phase. 
The authors of \cite{wang2017improving} used the output of a softmax function as a measure of confidence.  
%The output of a softmax function was used as a measure of confidence in \cite{wang2017improving}, 
However, a shortcoming of this approach is that a model could be uncertain in its predictions even with a high softmax output \cite{gal2016dropout}.

Bayesian neural networks \cite{neal2012bayesian} have been used to represent the uncertainty in the output of an NN. 
%Several variational inference methods have been introduced to deal with the prohibitive computational costs incurred during inference. 
%These methods, however, still come with a high computational cost. 
Two techniques that offset the high costs incurred during inference 
%which have been studied in the literature 
are bootstrapped ensemble NNs \cite{lakshminarayanan2017simple, osband2016deep} and dropout as Bayesian approximation \cite{gal2016dropout}. 
%The authors of \cite{osband2016deep} proposed \emph{bootstrapped DQN} as a means to accelerate the learning process. 
The bootstrap principle is to approximate the distribution of a population by a sample distribution \cite{efron1994intro}. 
This property allows us to use an ensemble of NNs to effectively represent uncertainty in the model where each network in the ensemble produces a value indicative of its confidence in its output. 
%Another approach to represent uncertainty in deep learning models was presented in \cite{gal2016dropout}, where the authors established an equivalence between \emph{dropout} training in deep NNs and Bayesian inference in Gaussian processes. 
In the dropout approach, an equivalence was established between \emph{dropout} training in deep NNs and Bayesian inference in Gaussian processes. 
We prefer the bootstrapped ensemble over dropout in this paper because of its lower time complexity during inference. 

%\textbf{Contributions}: 
\subsection{Contributions}
%In order to effectively represent model uncertainty and the confidence of the neural network in its output, we use an ensemble of neural networks with a shared network architecture. 

We use an ensemble of neural networks with a shared architecture to effectively represent human feedback and predict uncertainty in the model. 
%We use an ensemble of neural networks with a shared network architecture to effectively represent model uncertainty and the confidence of the NN in its output. 
The human feedback signal is then used as a shaping reward. 
We evaluate our method on the \emph{Bowling} and \emph{Skiing} Atari games in the Arcade Learning Environment \cite{bellemare13arcade}. 
We choose these games because many state-of-the-art deep reinforcement learning (DRL) algorithms are unable to achieve human-level performance. 
We observe that \emph{FRESH} is able to outperform state-of-the-art DRL algorithms in both environments. 
In the \emph{Bowling} game, \emph{FRESH} obtains a higher average score ($\mathbf{\sim 180}$) than a human expert ($\mathbf{\sim 150}$) \cite{mnih2015human}. 
%This is shown in Figure \ref{Bowling1}. 
In the \emph{Skiing} game, \emph{FRESH} performs as well as a human expert. 

%\begin{figure}
%	\includegraphics[width=2.7 in]{Figures/Experiments/Bowling_intro.pdf}
%	\caption{\emph{FRESH} outperforms state-of-the-art deep reinforcement learning algorithms, and also outperforms a human expert player in \emph{Bowling}.}\label{Bowling1}
%\end{figure}
%\textbf{talk about TAMER, Deep-TAMER- theirs is real valued human feedback; ours is binary feedback}. 
%, since it is typically easier for a human to judge whether an action is `good' or `bad' than associate a numerical value to the action. 

\subsection{Outline of Paper}

The paper is organized as follows: Section \ref{Sec:Preliminaries} gives a brief introduction to reinforcement learning. 
Our approach is outlined in Section \ref{Sec:Approach} and we present detailed results of our experiments in Section \ref{Sec:ExperimentalEvaluation}. 
Section \ref{Sec:RelatedWork} summarizes related work and we conclude in Section \ref{Sec:Conclusion}.

\section{Reinforcement Learning}\label{Sec:Preliminaries}
%
%\subsection{Reinforcement Learning}

A Markov decision process (MDP) \cite{puterman2014markov} is a tuple $(S,A,\mathbb{T},\rho_0, R)$. 
$S$ is the set of states, $A$ the set of actions. 
$\mathbb{T}:S \times A \times S \rightarrow [0,1]$ encodes $\mathbb{P}(s_{t+1}|s_t,a_t)$, the probability of transition to $s_{t+1}$, given current state $s_t$ and action $a_t$. 
$\rho_0$ is a probability distribution over the initial states. 
$R : S \times A \times S \rightarrow \mathbb{R}$ denotes the reward that the agent receives when transitioning to state $s_{t+1}$ from $s_t$ while taking action $a_t$. 
%In this paper, $R < \infty$. 

The goal for an RL agent \cite{sutton2018reinforcement} is to interact with its environment (that is modeled as an MDP) and learn a \emph{policy} $\pi$, 
%which is a map from states to actions 
in order to maximize 
%the expected discounted cumulative reward, i.e. $\max~$ 
$J:=\mathbb{E}_{\tau \sim \pi}[\sum_{t=0}^{\infty}\gamma^t R(s_t,a_t, s_{t+1})]$. 
Here, $\gamma$ is a discounting factor, and the expectation is taken over the trajectory $\tau=(s_0,a_0,r_0,s_1,\dots)$ induced by policy $\pi$. %$\pi_{\theta}$. 
If $\pi: S \rightarrow A$, the policy is \emph{deterministic}. 
On the other hand, a randomized policy returns a probability distribution over the set of actions, and is denoted $\pi: S \times A \rightarrow [0,1]$. 
It should be noted that the agent does not have direct access to the transition probabilities or the reward. 

The value of a state-action pair $(s,a)$ following policy $\pi$ is represented by the \emph{Q-function}, $Q^{\pi}(s,a) = \mathbb{E}_{\tau \sim \pi}[\sum_{t=0}^{\infty}\gamma^t R(s_t,a_t, s_{t+1})|s_0=s,a_0=a]$. %, which is the expected discounted cumulative return from state-action pair $(s,a)$.
The Q-function allows us to calculate the state value $V^{\pi}(s) = \mathbb{E}_{a \sim \pi}[Q^{\pi}(s,a)]$. 
The advantage of a particular action $a$, over other actions at a state $s$ is defined by $A^{\pi}(s,a) := Q^{\pi}(s,a)-V^{\pi}(s)$. 
The overall policy can be determined by (dynamically) updating Q-values, using Q-learning \cite{watkins1992q}. 
Exploration strategies in the environment at the early stages of learning include $\epsilon$-greedy (choose action that maximizes $Q$-value with high probability, and a random action with low probability) and Boltzmann exploration (choose action according to a probabilistic model).

\section{Feedback-based Reward Shaping}\label{Sec:Approach}
The use of human feedback in high-dimensional state spaces raises the following questions:
\begin{enumerate}
\item[\textbf{Q1:}] How can feedback signals provided by a human operator be effectively used in high-dimensional state spaces?

\item[\textbf{Q2:}] How can this feedback be meaningfully integrated with deep reinforcement learning (DRL) algorithms?
\end{enumerate}
In this section, we present \textbf{\emph{FRESH}}, a feedback-based reward shaping approach towards answering these questions. 
We assume that an RL agent has to learn to maximize its reward in an environment that has a high-dimensional state space. 
An example of such an environment is an Atari game in the Arcade Learning Environment. 
`States' in this environment are collections of pixels (images). 
Although deep NNs have been successful in outperforming a human expert player in many Atari games \cite{mnih2015human}, there are some games where the expert is still able to perform better than state-of-the-art DRL algorithms. 
This paper studies two examples of the latter (Bowling and Skiing), and we demonstrate that \emph{FRESH} performs at least as well as a human expert player in these environments. 

%In this work, we tackle two problems: (1) how to effectively use binary human feedback in high-dimensional state spaces, and (2) how to combine human feedback and deep reinforcement learning algorithms. 
%In this section, we introduce how we deal with the two issues. 
%First we introduce how we make use of binary human feedback by training a deep neural networks, and then the method of combining output of feedback networks into reinforcement learning is presented. 
\subsection{Binary-valued Human Feedback}
We assume that a human operator is able to provide binary-valued feedback on actions and states. 
We do not require the operator to be an expert. 
However, we assume that this operator will have the ability to understand game-play in the environment after explanation by an expert. 
%, but have the ability to understand the environment after an explanation by an expert...(\textbf{think and write}). 
Feedback given on an action is a (local) interpretation of the quality of the action at a particular state, independent of how `good' or `bad' the state is. 
In comparison, feedback provided on a state is a (more global) interpretation on whether the current observed state is good or bad %relative to the sequence of states and actions preceding it. 
%This latter feedback will also depend on whether the state in question is `good' in the overall environment, 
in terms of the rewards that can be obtained. 
%(\textbf{IS THIS LAST SENTENCE CORRECT?}). % \textcolor{blue}{This is not exactly correct. Actually, when we evaluate a state, we are not just comparing the state with the states preceding it. We also have some sort of idea about what states should be good for the whole game. Note that we first visualize a trajectory to human, and then obtain feedback for each state/action in the trajectory.}
The reason we use this type of feedback signal is that in many settings, rather than assigning a numeric value, it is relatively easier for a human to interpret whether a state or an action taken in a particular state is simply `good' or `bad'. 
%
%The feedback on an action is a local point of view from human on the quality of action in a specific state no matter how bad or good the state is.
%In contrast, the feedback on a state is a more global viewpoint of human on whether the state is prefered in the task.
%The motivation why we focous on binary feedback is that, for many senarios, it is much easier for a human expert to tell if a state is good or bad, rather than give a numerical value on how good or how bad the state is. 
%The argument is the same for giving comments on actions.
%More precisely, 
Specifically, the human provides feedback on actions implicitly according to %certain 
a hidden function $h_{s,a}(\cdot,\cdot):S \times A \rightarrow \{0,1\}$, and feedback on states according to a hidden function $h_s(\cdot):S \rightarrow \{0,1\}$.
In each case, $0/1$ denote \textit{bad}/ \textit{good}.  
The human operator can additionally provide feedback signals that indicate \emph{not sure} (if they are not certain whether an action in a state is `good' or `bad') and \emph{skip} (if the operator feels that sufficient number of feedback signals have been provided for a trajectory). 
These latter two signals are not used to train the agent, but they allow the human operator to significantly reduce the amount of time spent in the training phase. 
%are able to significantly reduce the amount of time spent by the human operator during training. 
%achieve a significant reduction in the amount of time spent by the human operator. 
We note that both $h_{s,a}$ and $h_s$ can be time-varying, implying that the feedback can change for the same state or state-action pair as the agent training process proceeds. 
Moreover, the human operator might also change their expectations of the agents' performance over time, which justifies the time-varying nature of the feedback signal. 
%
%Different from tabular cases, in testing phase many states may never have been seen during training phase in environments with high-dimensional state spaces. 
%In addition, the number of human feedback is limited. 
%Therefore we need to generalize from the human feedback for states and actions which are in some sense similar to current ones. 
%%Here we assume that similar states and actions should have similar human feedback. 
%To take advantage of human feedback, we resort to the generalization capability of deep neural networks (NN).

We assume that at each time, in a state $s_t$, there is exactly one action $a_t$ that is the `best' in that state. 
That is, taking this action will lead to the agent receiving a higher (accumulated) reward than taking any other action. 
We formulate a binary classification problem, and use maximum likelihood estimation (MLE) to determine the best action in a state. 
%
%For the binary feedback on actions, we formulate a classification problem under the assumption that only one action is the best in a specific state.
%That is, given a state, the NN needs to figure out which action in the action space is the best in that state.
%To achieve this, we use maximum likelihood estimation (MLE), such that the NN prediction maximizes the probablity of binary human feedback.
We denote the predicted probability that action $a_i$ is the best in state $s$ by $f^{a_i}(s)$. Furthermore, $\sum_{i=1}^{|A|}f^{a_i}(s)=1$.
To implement our maximum likelihood estimator, we use cross-entropy loss with both positive and negative labels:
\begin{align}%\label{loss_a}
\mathcal{L}(f^{a_i}(s), h_{s,a_i}) &= -h_{s,a_i}\log f^{a_i}(s) - (1-h_{s,a_i})\log\sum_{a\neq a_i}f^{a}(s), \nonumber
\end{align}
where $h_{s,a_i}$ is the binary-valued human feedback associated to state-action pair $(s,a_i)$.

We formulate an analogous classification problem for feedback on states. %, we formulate a different classification problem.
Let $g(s)$ denote the predicted probability that state $s$ is \textit{good}. %} and $h_s$ denote the binary human feedback for state $s$. 
The loss function used to evaluate this prediction is:
\begin{align}%\label{loss_s}
\mathcal{L}(g(s), h_{s}) = -h_{s}\log g(s) - (1-h_{s})\log (1-g(s)),\nonumber
\end{align}
%During inference phase, given a state $s$, the prediction becomes $\arg\max_{a_i}f^{a_i}(s)$

\subsection{Generalizing Human Feedback}
A challenge in high-dimensional state spaces that is not encountered in the tabular setting is that the agent might observe a lot of states at test-time that it might have never seen during training. 
Moreover, since the number of feedback signals that a human can provide is often limited, it is important to be able to \emph{generalize} from the feedback in order to determine whether a state encountered at test time is similar to some state seen during training. 
In order to fully exploit the feedback signals given by a human, we leverage deep neural networks (DNNs). DNNs have been shown to have the ability to use feedback on states and actions and generalize it to other states and actions that have not been previously seen, but `similar' to those already known \cite{goodfellow2016deep}. 
We term the neural networks used to abstract the feedback provided by a human operator as \emph{feedback neural networks (FNNs)} for the rest of this paper.
%\textcolor{blue}{Abstract feedback on states/ actions and generalize it to other states/ actions which are unseen before but similar to the current ones.}

Using NNs to generalize human feedback presents another challenge. 
NNs typically produce an output for any input, and do not have a measure of confidence. 
%Simple (\textbf{WHAT IS simple here?}) neural networks always produce an output for any input. \textcolor{blue}{Change to "Neural networks generally produce an output for any input and do not have a measure of uncertainty" ?}
This could lead to a scenario when the network produces an arbitrary output to an input state that it has never seen during training, and this could lead the RL agent to undesired/ incorrect states. 
This necessitates development of a mechanism that allows the agent to reject the output of FNNs. 

A class of NNs called \emph{Bayesian neural networks} \cite{neal2012bayesian} are able to produce both an output, and a value indicating the confidence of the NN in the output. 
However, this process is computationally expensive. 
To obtain these confidence values, approximations of Bayesians NNs using dropout \cite{gal2016dropout} and 
ensemble of NNs for predicting model uncertainty \cite{lakshminarayanan2017simple, osband2016deep} have been proposed. 
%combining exploration of the environment with DNNs via bootstrapping an ensemble of NNs \cite{osband2016deep} have been recently proposed. 
%
%One challenge of applying neural networks for generalizing human feedback is that simple neural networks always give output whatever the input is.
%If the neural networks give an arbitrary output when input a state which is never seen and very different from states in training phase, then the agent may be mislead to some undesired point.
%In order to alleviate that issue, we need a certain mechanism that allows the agent to reject the output of feedback neural networks (FNN).
%To achieve this, the FNN should be able to output a confidence value indicating how confident it is regarding its output.
%Baysian neural networks are able to produce both output and the confidence, but it is computational cost is prohibitive. 
%To obtain model uncertainty, approximated Baysian method for neural networks, such as droput as a bayesian approximation and bootstrap esemble neural networks, was proposed recently.
%In this work, 
We use the bootstrap ensemble NN architecture in this paper due to its lower time complexity during inference when compared to dropout.
\begin{figure}
	\includegraphics[width=2.0 in]{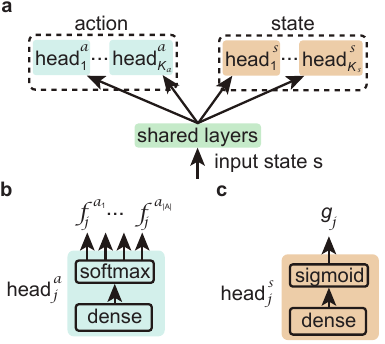}
	\caption{Figure \textbf{a} shows the multi-head architecture with shared layers for the feedback neural network that we use in this paper. Figures \textbf{b} and \textbf{c} show architectures of parts of the networks corresponding to learning feedback on actions and feedback on states respectively.}\label{networks}
\end{figure}

In order to achieve a further reduction in complexity, we use a shared network architecture as shown in Figure \ref{networks}. 
In this architecture, the entire network has $K_s+K_a$ heads, with $K_s$ bootstrapped heads for learning feedback on states and $K_a$ bootstrapped heads for learning feedback on actions, along with a shared network. 
Each head is randomly initialized and trained on a bootstrapped subset of feedback data, that is sampled with replacement from the entire feedback data. 
All heads share the same preceding layers, which allows the entire network to be trained more efficiently. % adn the whole networks can be trained more effeciently.
We collect the prediction from each head and use the mean of these predictions as the final prediction. 
%We use the mean of the predicted distribution as the final prediction during inference. (\textbf{this sentence needs to be more clear}) \textcolor{blue}{We collect the prediction from each head and calculate the mean as the final prediction.}
%For inference, we calculate the mean of predicted distribution as the final prediction. 

Let $\mathbf{f}_j=\big(f^{a_1}_j(s), f^{a_2}_j(s),\dots,f^{a_{|A|}}_j(s)\big)$ denote the output of the $j$-th action head, where $f^{a_i}_j(s)$ is the probability that action $a_i$ is the best in state $s$, and let $g_j(s)$ denote the output of the $j$-th state head. 
In order to predict the human feedback on actions, we define:
%\begin{align}\label{pred_action}
%\text{pred}_\text{action}(s)=\arg \max_{a_i}\frac{1}{k_a}\sum_{j=1}^{k_a}\mathbf{f}_j(s)
%\end{align}
\begin{align}%\label{pred_action}
\text{pred}_\text{action}(s)= \argmax_{i\in \{1,\dots,|A|\}}\big\{\frac{1}{K_a}\sum_{j=1}^{K_a}f^{a_i}_j(s)\big\},\nonumber
\end{align}
and to predict human feedback on states, we define:
\begin{align}%\label{pred_action}
\text{pred}_\text{state}(s)= \begin{cases}
1, & \text{if }
\begin{aligned}[t]
\frac{1}{K_s}\sum_{j=1}^{K_s}g_j(s)>0.5,
\end{aligned}
\\
0, & \text{otherwise.}
\end{cases}\nonumber
\end{align}
We calculate the empirical standard deviation of individual prediction by different heads and use it as a proxy for the confidence value.
Let $p^a_j(s)=\argmax \limits_{i\in \{1,\dots,|A|\}}f^{a_i}_j(s)$ denote the prediction of the $j$-th action head. Then, the confidence value $c_a(s)$ for predicting human feedback on actions can be obtained by computing the empirical standard deviation:
\begin{align}
c_a(s) = 1- \sqrt{\frac{1}{K_a-1}\sum_{j=1}^{K_a}\bigg(p^a_j(s)-\frac{\sum_{j=1}^{K_a}p^a_j(s)}{K_a}\bigg)^2}\nonumber
\end{align}
The (proxy for the) confidence $c_s(s)$ of predicting the human feedback on states can be computed similarly. 

\subsection{Reward Shaping}

The reward signal supplied by the environment can often be sparse and/ or significantly delayed, although it can \textit{accurately} define desired goals for the agent. 
%be \textit{flawless} (\textbf{this is a vague term, and needs to be precisely defined if we're using it}) \textcolor{blue}{we can remove "flawless", and just say "accurately defines the tasks"}. 
%For many scenarios, human expert is able to comment on whether a state and an action is good or not.
In order to make the learning procedure more efficient, the human feedback can be introduced through reward shaping.
Once an approximation of the human feedback model is available, we can transfer the model output to more frequent and timely reward signals, although this additional reward may sometimes be incorrect due to an error made by human operator providing the feedback. 

In order to make use of the feedback given by the human operator, and not limit the agent's performance when this operator makes an error, we present a way to combine FNNs with the environment reward. 
%In this section, how we combine FNN and environment reward will be introduced.
Let $s_t$, $a_t$, and $s_{t+1}$ respectively denote the state and action at time $t$, and the next state after taking action $a_t$ in $s_t$. 
The feedback received on actions is transferred to an additional reward $r_a$ as:
\begin{align}%\label{shaping_a}
r_a(s_t,a_t,s_{t+1}) = \begin{cases}
1, & \text{if }
\begin{aligned}[t]
a_t = \text{pred}_\text{action}(s_t)~~ \text{and}~~c_a(s_t)>1-\beta_a
\end{aligned}
\\
0, & \text{otherwise,}
\end{cases}\nonumber 
\end{align}
where $\beta_a$ is a pre-assigned constant threshold. 
If the action taken by the agent is consistent with the best action predicted by the FNN and the confidence of FNN in the prediction $c_a(s_t)$ is higher than $1-\beta_a$, then an additional reward is provided.
Similarly, feedback received on states can be transferred to an additional reward $r_s$ as:
\begin{align}%\label{shaping_s}
r_s(s_t,a_t,s_{t+1}) = \begin{cases}
1, & \text{if }
\begin{aligned}[t]
\text{pred}_\text{state}(s_{t+1})=1~~ \text{and}~~c_s(s_{t+1})>1-\beta_s
\end{aligned}
\\
0, & \text{otherwise,}
\end{cases} \nonumber 
\end{align}
If $r_e$ denotes the environment reward, the \emph{shaped reward} $r(s_t,a_t,s_{t+1})$ is then:
\begin{align}\label{shaped_r}
r(s_t,a_t,s_{t+1}) = r_e(s_t,a_t,s_{t+1})+\lambda_a r_a(s_t,a_t,s_{t+1})+\lambda_s r_s(s_t,a_t,s_{t+1}),
\end{align}
where $\lambda_a$ and $\lambda_s$ are the weights, balancing the importance of the three rewards.
Although $\lambda_a$ and $\lambda_s$ can decay over time,  during our experiments we observed that keeping them constant works well. 
%
%\textbf{Can we use a notation different from $h_a$ and $h_s$ for the thresholds? This is because the same notation is used in Sec 4.1 for the hidden functions} \textcolor{blue}{Has changed them to $\beta_a$ and $\beta_s$.}

\subsection{Algorithm}

We evaluate \emph{FRESH} when it is used with deep RL algorithms that use a replay buffer for learning, e.g. DQN. 
%
%In this work, we consider applying interactive reward shaping to deep reinforcement learning algorithms which utilizes replay buffer for learning, e.g., DQN.
%
\begin{algorithm}
	\caption{HumanFeedbackCollection}
	\begin{algorithmic}[1] \label{Algo1}
		\renewcommand{\algorithmicrequire}{\textbf{Input:}}
		\REQUIRE Replay buffer $B_q$ storing trajectory experience and buffer $B_f$ storing human feedback. Masking distributions $M_s, M_a$
%		\textbf{Initialization}: \\
		\STATE Sample a trajectory $\tau$ from $B_q$ and visualize $\tau$ for feedback.
		\FOR{$(s_t,a_t,r_t,s_{t+1}) \in \tau$}
		\IF{new feedback on state $f_s$ is available}
		\STATE sample $\mathbf{m}_s \sim M_s$ and store $(s_{t+1}, f_s, \mathbf{m}_s)$in $B_f$
		\ENDIF
		\IF{new feedback on action $f_a$ is available}
		\STATE sample $\mathbf{m}_a \sim M_a$ and store $(s_t,a_t,f_a,\mathbf{m}_a)$ in $B_f$
		\ENDIF
		\ENDFOR
		\RETURN{$B_f$}
	\end{algorithmic}
\end{algorithm}
The procedure for collecting the human feedback is summarized in Algorithm \ref{Algo1}.
First, we sample a trajectory $\tau$ from the replay buffer $B_q$ (line 1).
Any choice of sampling method can be used in this step. 
%Different sampling method can be adopted in this step. 
For example, the trajectory with highest or lowest reward may be given higher priority at different training stages. 
This sampled trajectory is then presented to the human operator in order to receive feedback. 
The speed at which $\tau$ is displayed to the human can be much slower than actual game play. 
%Once a trajectory $\tau$ is sampled, it will be visualized for feedback inquiry. 
%The visualization speed can be much slower than real-time play.
For example, when the states are represented by images, we can apply a lower frame rate so that it will be easier for the operator to assess states and actions in the trajectory. 
After feedback on a state $f_s$ or feedback on an action $f_a$ is provided, a mask $\mathbf{m}_s \in \mathbb{Z}^{K_s}$ or $\mathbf{m}_a \in \mathbb{Z}^{K_a}$ will be sampled from a masking distribution. This will indicate which heads this feedback should be used to train.
For example, the components of the mask can be independently drawn from a Bernoulli distribution (double or nothing bootstrap) or from an exponential distribution (Bayesian non-parametric posterior of a Dirichlet process) \cite{osband2016deep}.
We note that feedback will not be provided on all states or actions, since the human operator might refuse to provide an assessment if they are not sure of the quality of the state/action.
The feedback together with the mask will be stored in the feedback buffer (line 3-8).
%

%A full description of deep interactive reward shaping is given in Algorithm \ref{Algo2}. 
Algorithm \ref{Algo2} describes \emph{FRESH}. 
The feedback buffer is initialized by providing feedback on trajectories from random play to train the feedback networks FNN (lines 1-7). The FNNs are updated using stochastic gradient descent (SGD). 
For training the value networks $Q$, we use a DQN-based algorithm \cite{van2016deep, wang2016dueling}, but the reward function is changed to Equation (\ref{shaped_r}) (lines 9-12). 
The human operator is asked to provide feedback every $N_c$ episodes (lines 13-15). 
%Every $N_c$ episodes, human will be inquired for feedback (line 13-15).
If $N_f$ new feedback signals are available, the FNN is re-tuned (lines 16-19). 
\begin{algorithm}
	\caption{\emph{FRESH} for DQN}
	\begin{algorithmic}[1] \label{Algo2}
		\renewcommand{\algorithmicrequire}{\textbf{Input:}}
		\REQUIRE Value networks $Q$. Feedback networks FNN. Masking distributions $M_s, M_a$. Replay buffer $B_q$ storing experience for training DQN and buffer $B_f$ storing human feedback for training FNN. Thresholds $\beta_a$ and $\beta_s$. Weights $\lambda_a$ and $\lambda_s$. Feedback collection frequency $N_c$ and FNN update frequency $N_f$. Initial number of trajectories and feedback $n_i$ and $m_i$.
		
%		\hspace{7.5mm}Differentiable value function $V(s;\omega)$ parametrized by $\omega$
%		
%		\hspace{7.5mm}Illustrative rule of environment $f(s_t,a_t)$
%		
%		\hspace{7.5mm}$T_{max}$, $t_{max}$
		
%		\textbf{Initialization}: \\
		\REPEAT
			\STATE Sample trajectories based on random play and store trajectories in $B_q$.\\
		\UNTIL{Collect $n_i$ trajectories in $B_q$}
		\REPEAT
			\STATE $B_f = HumanFeedbackCollection(B_q,B_f,M_s,M_a)$
		\UNTIL{collect $m_i$ feedback in $B_f$}
		\STATE sample batches from $B_f$ and update FNN using SGD
		\FOR{Episode i=1, $\dots$}
			\REPEAT
				\STATE execute action $a=\argmax_a Q(s,a)$, observe reward $r$ and next state $s'$ and store $(s,a,r,s')$ in $B_q$
				\STATE update $Q$ using DQN algorithm but change reward function to Equation (\ref{shaped_r})
			\UNTIL{end of episode}
			\IF{i \% $N_c==0$}
				\STATE $B_f = HumanFeedbackCollection(B_q,B_f,M_s,M_a)$
			\ENDIF
			\IF{Number of new feedback $new_f>N_f$}
				\STATE sample batches from $B_f$ and update FNN using SGD
				\STATE $new_f\leftarrow0$
			\ENDIF
		\ENDFOR
	\end{algorithmic}
\end{algorithm}

\section{Experimental Evaluation}\label{Sec:ExperimentalEvaluation}

\begin{figure}
	%	\centering
	\begin{minipage}{.2\textwidth}
		\centering
		\includegraphics[width=.63\linewidth]{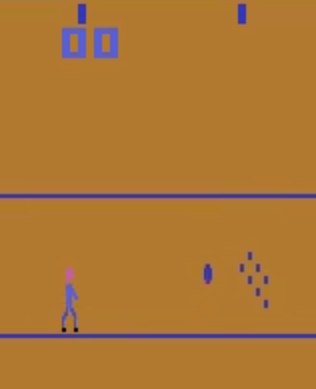}
		%		\captionof{figure}{Skiing}
		%		\label{Skiing}
	\end{minipage}%
	\begin{minipage}{.2\textwidth}
		\centering
		\includegraphics[width=.63\linewidth]{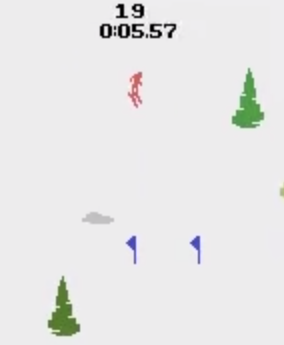}
		%		\captionof{figure}{Bowling}
		%		\label{Bowling}
	\end{minipage}
	\caption{Snapshots from game-play in the Atari games of Bowling (left) and Skiing (right). In Bowling, the goal for the player is to roll the ball and knock down as many pins as possible. In Skiing, the goal for the player is to reach the bottom of a valley as soon as possible, and at the same time, pass through as many gates as possible while avoiding obstacles.}
	\label{Env}
\end{figure}

We evaluate \emph{FRESH} on two Atari games in the Arcade Learning Environment \cite{bellemare13arcade}. 
Figure \ref{Env} shows screen shots of game play in the \emph{Bowling} and \emph{Skiing}  environments. 
Although human experts can achieve high scores with relative ease in these games, it has been extremely difficult for state-of-the-art deep reinforcement learning (DRL) algorithms to match this. 
We observe that the performance on these games using \emph{FRESH} compares favorably with that of a human expert, and is significantly improved over other DRL algorithms.

\subsection{Game Description and Experiment Setup}
%Screenshots of Atari game of Bowling and Skiing are shown in figure \ref{Env}.
The Bowling game comprises four actions, \textit{no-action, up, down, fire}. 
A game lasts 10 rounds and the player (agent) gets two chances in each round to roll the bowling ball to knock down as many pins as possible. 
The game begins with the player choosing a position to release (\textit{fire}) the ball by moving vertically using actions \textit{up} or \textit{down}.
After releasing the ball, the player gets one chance to steer the direction of the ball by taking actions \textit{up} or \textit{down}. 
The reward structure of the game makes it difficult for state-of-the-art deep reinforcement learning (DRL) algorithms to obtain a high reward. 
In particular, the player does not receive an immediate reward if all pins are knocked down in one turn. 
Instead, this reward will be supplied at the end of the next turn, together with the reward for that turn. 
This makes it difficult for DRL algorithms to correctly provide credit for actions by simply looking at the reward. 

The Skiing game consists of three actions, \textit{no-action, left, right}. 
The player controls their direction of motion in order to avoid obstacles (trees and moguls) and pass through the gates. 
The goal for the player is to reach the bottom of the valley as soon as possible, and in the process, pass through as many gates as possible.
This game is difficult for DRL algorithms to play since the reward indicating the number of gates passed through is supplied only at the end the game, making the credit assignment task difficult.

%There are four actions in Bowling: \textit{no-action, up, down, fire} and three actions in Skiing: \textit{no-action, left, right}.
%In bowling, the game lasts 10 frames and the player (agent) has 2 chances in each frame to roll the bowling ball in order to knock down as many bowling pins as possible. 
%The player begins with choosing a position to release (\textit{fire}) the ball moving vertically using action \textit{up} or \textit{down}.
%After releasing the ball, the player has one chance to steer the direction of the ball by taking action \textit{up} or \textit{down}.
%Bowling is difficult for state-of-the-art deep reinforcement learning (DRL) algorithms due to its reward structure. 
%The player will not be rewarded immediately if all pins are knocked down in one turn, but instead the reward will be given together with the reward received in the next turn.
%Therefore, it is hard for DRL algorithms to correctly credit the actions by just looking at the reward. 
%In Skiing, the player control the moving direction in order to avoid obstacles (trees and moguls) and pass through the gates. 
%The goal in the skiing is to reach the bottom of the valley as rapidly as possible, and in the meanwhile pass through as many gates as possible.
%Skiing is difficult for DRL algorithms since the reward indicating how many gates has been passed through is given at the end the game.
%The extremely delayed reward makes the credit assignment task hard.

In all our experiments, we use the same neural network architecture and hyper-parameters for DQN as \cite{wang2016dueling} does.
Additionally, we apply double Q-learning \cite{van2016deep} to avoid overestimation of action values.
The shared network of our feedback neural network (FNN) has the same architecture as the convolutional layers of DQN, and each head of  the FNN adopts three fully-connected layers with batch-normalization. 
Each state is a tensor stacked by 4 gray-scale images, obtained by converting 4 consecutive colored video game frames. 
The regions of the frames showing the game score are removed when the frame is an input to the FNN. 
%\textcolor{blue}{For the input of FNN, we remove the score regions from the frames.}
We clip the environment reward using the same approach as \cite{mnih2015human} does, and set $\lambda_a=0.2$ and $\lambda_s=0.1$ across the experiments. 
In the early stages of training the FNN, when sufficient diverse feedback signals are not available, the agent might be distracted from the true goal and run into a \emph{cycling} problem, as indicated in \cite{Ng1999policy}. 
To alleviate this problem, when $|s_t-s_{t-1}|$ is smaller than a known threshold value (which is indicative of the agent being stuck in a local state), we give the agent a penalty to offset the reward supplied by the FNN. 
This works well in our experiments. 
%\textcolor{blue}{When the number of diversified feedback for training FNN is not enough in the early training stage, the agent may run into 'cycling' issue \cite{Ng1999policy}. To alleviate the problem, we give the agent a penalty to offset the reward by FNN when $|s_t-s_{t-1}|$ is smaller than a given threshold, which is an indication that the agent is stucked in certain local states. This works well in our experiments.} 
Bernoulli distribution is used as the masking distribution.
To evaluate Algorithm \ref{Algo2}, a human operator provides feedback using a computer with an user interface to visualize trajectories and receiving feedback. 
While providing feedback, the human trainer is allowed to say \emph{not sure} for states and actions for which the human is not sure about the quality. 
In the following, we first evaluate the performance of \emph{FRESH} in both Bowling and Skiing and then present a detailed study on the effect of the different components of the FNN in Bowling.

\subsection{\emph{FRESH} for Bowling and Skiing}  

We first collect $n_i=100$ trajectories using random play, which are sorted based on the accumulated reward. The trajectory with the highest reward is given the highest priority to obtain human feedback. 
%This will allow us to balance the proportions of \textit{good} states/actions and \textit{bad} ones to some extent. 
%In this way, we can balance the number of \textit{good} states/actions and the number of \textit{bad} ones to some extent.
%\textcolor{blue}{By looking at trajectories from random play}, we collect $m_i=500$ feedback signals for Bowling and $m_i=2000$ signals for Skiing. 
We collect a total of $m_i=500$ feedback signals for Bowling and $m_i=5000$ signals for Skiing from trajectories of random play.  
While training the DQN, we collect feedback signals every $N_c=30$ episodes. 
When $N_f=300$ new feedback signals is available, we update the FNN using all the data in the feedback buffer. 
%\textcolor{red}{When calculating the number of feedback signals, we do not distinguish between feedback on actions and feedback on states. 
%In our experiments, the number of feedback signals provided on states is approximately equal to the number of feedback signals on actions.}
In our experiments, we observed that the ratio of the feedback signal \textit{good} over \textit{bad} is around 0.2.
In this section, we fix the number of heads $K_a=K_s=10$ and thresholds $(\beta_a=1.0,\beta_s=0.02)$.

\begin{figure}
	\includegraphics[width=2.5 in]{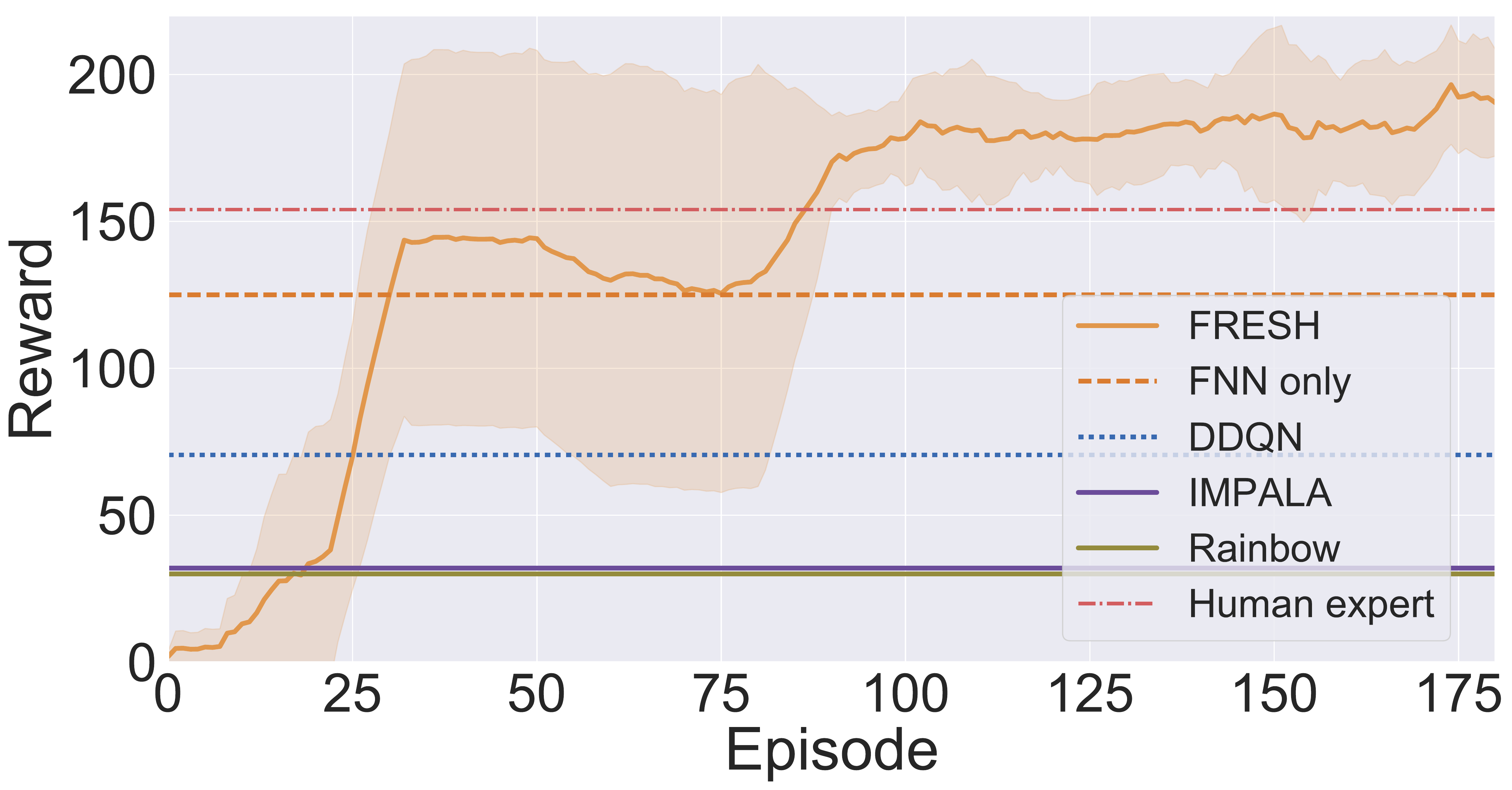}
	\caption{\emph{FRESH} outperforms state-of-the-art deep reinforcement learning algorithms, and also outperforms a human expert player in Bowling. The shaded region indicates the variance of the reward.}\label{Bowling}
\end{figure}

\begin{figure}
	\includegraphics[width=2.5 in]{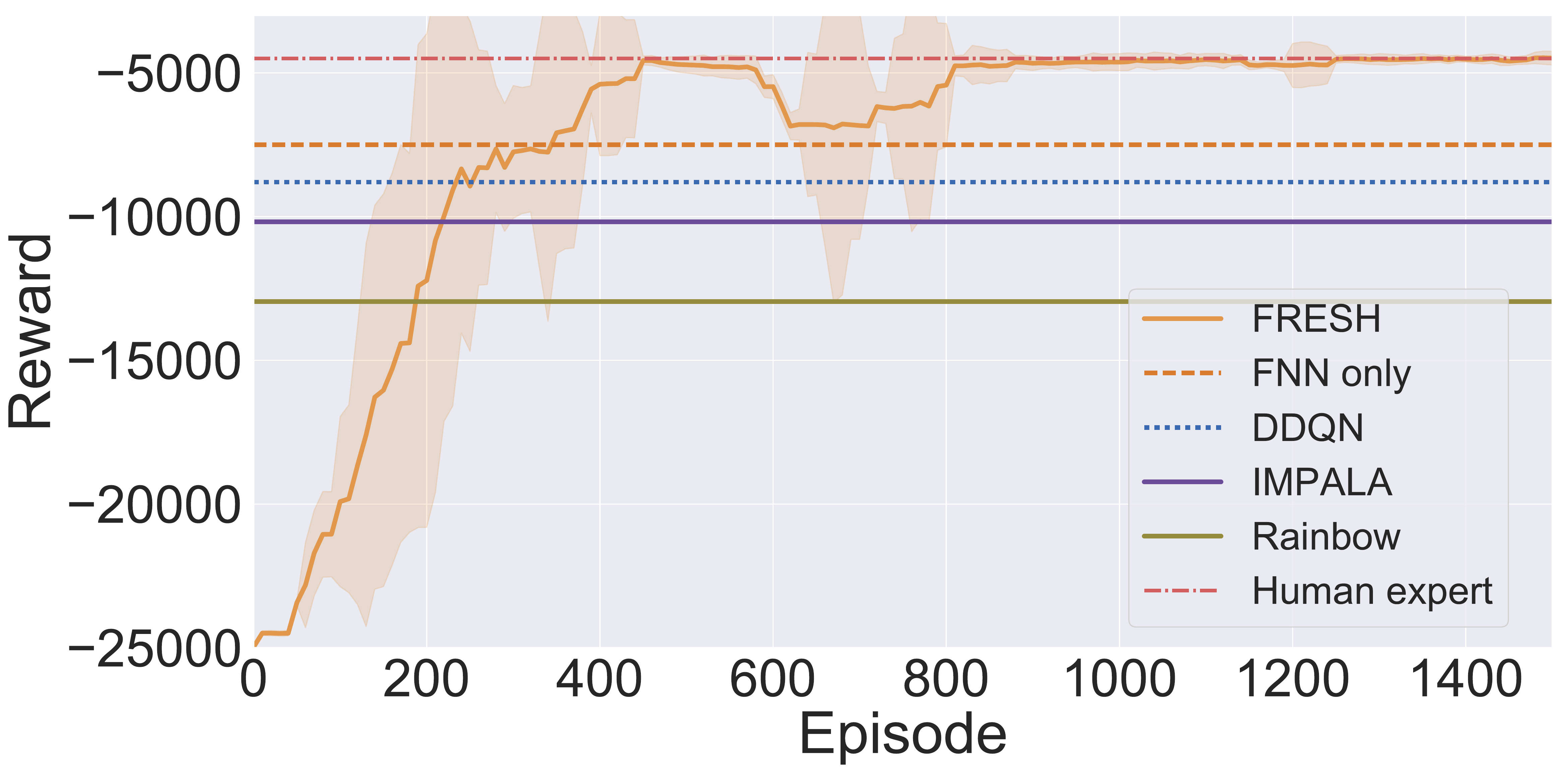}
	\caption{\emph{FRESH} outperforms state-of-the-art deep reinforcement learning algorithms, and performs as well as a human expert player in Skiing. The shaded region indicates the variance of the reward.}\label{Skiing}
\end{figure}

Figures \ref{Bowling} and \ref{Skiing} show the performance of \emph{FRESH} in \emph{Bowling} and \emph{Skiing}.
Each curve is an average over five runs with different human feedback and initialization.
Shaded regions indicate the standard deviation.
We compare the performance of \emph{FRESH} (Algorithm \ref{Algo2}) with Double-DQN \cite{van2016deep}, IMPALA \cite{espeholt2018impala}, Rainbow \cite{hessel2018rainbow}, and Human expert.
We use the best final performance reported in the literature for Double-DQN, IMPALA and Rainbow.
The human expert performance data is from \cite{mnih2015human}.
Since the trained FNN is able to output an estimated best action for every state, it can also be used as a policy. 
Therefore, we report the performance of using FNN alone as well. 

\emph{FRESH} allows the agent to learn good policies in both environments. In contrast, state-of-the-art DRL algorithms fail even after training for $>100$ million frames. 
\emph{FRESH} lets the agent learn policies that allow it to obtain higher scores than a human expert in \emph{Bowling}. 
This indicates that \emph{FRESH} can not only guide the learning process but also that the final performance is not limited by the quality of feedback. 
In \emph{Skiing}, \emph{FRESH} outperforms the policy induced by the FNN and achieve a similar score as a human expert.

We observe that \emph{FRESH} allows the agent to achieve an average score of $\mathbf{187}$ in \emph{Bowling}. 
This is $\mathbf{21.4 \%}$ higher than the average score obtained by a human expert in this environment. 
The highest score obtained by our algorithm in \emph{Bowling} is $\mathbf{> 200}$. 
State-of-the-art deep RL algorithms perform poorly in comparison- DDQN obtains an average score of 70.5, while the other methods report an even lower score. 
Deep-TAMER \cite{warnell2018deep}, on the other hand, is able to achieve an average score of around $180$, and a high-score of around $200$. 
In \emph{Skiing}, \emph{FRESH} is able to obtain an average score of $\mathbf{-4400}$, which is equal to that obtained by an human expert player. 
The deep RL algorithms perform poorly in this environment also. The authors of \cite{warnell2018deep} do not report a score in the Skiing environment.

\subsection{Discussion: Bowling}

In order to better understand the performance of our algorithm, we carry out several modifications:
\begin{enumerate}
\item \emph{Type of feedback}: We train the FNN using feedback on actions only, feedback on states only, or feedback on both actions and states. The number of heads of the FNN is fixed at $K_a=K_s=10$, and the thresholds are fixed to $(\beta_a=1.0,\beta_s=0.02)$.
\item \emph{Number of heads}: We modify the number of heads of the FNN. We consider the cases: $K_a=K_s=1$ (no ensemble), $K_a=K_s=5$, $K_a=K_s=10$ and $K_a=K_s=20$. When $K_a, K_s >1$, thresholds are fixed to $\beta_a=1.0$ and $\beta_s=0.02$. Feedback on both states and actions is used to train the FNN.
\item \emph{Thresholds}: We fix $K_a=K_s=10$ and use feedback on both states and actions to train the FNN, but vary the thresholds. We compare: no threshold, $(\beta_a=1.0,\beta_s=0.02)$, $(\beta_a=1.0,\beta_s=0.2)$, $(\beta_a=0.5,\beta_s=0.02)$.
\end{enumerate}
Due to the difficulty of obtaining large amounts of human interaction data, we reuse feedback data collected from the experiments of the previous section and train FNN at the start of DQN training.
%In this way, we can also eliminate the uncertainty factor introduced by human and show the effect of each element of FNN to the maximum extent. (\textbf{this sentence needs to be more clear})\textcolor{blue}{
%\textcolor{red}{Reusing the feedback data allows us to alleviate the uncertainty introduced by the human operator when providing the feedback signal. In this way, we can focus on analyzing the effect of each modification to the FNN enumerated above.}

\begin{figure}
	\includegraphics[width=2.5 in]{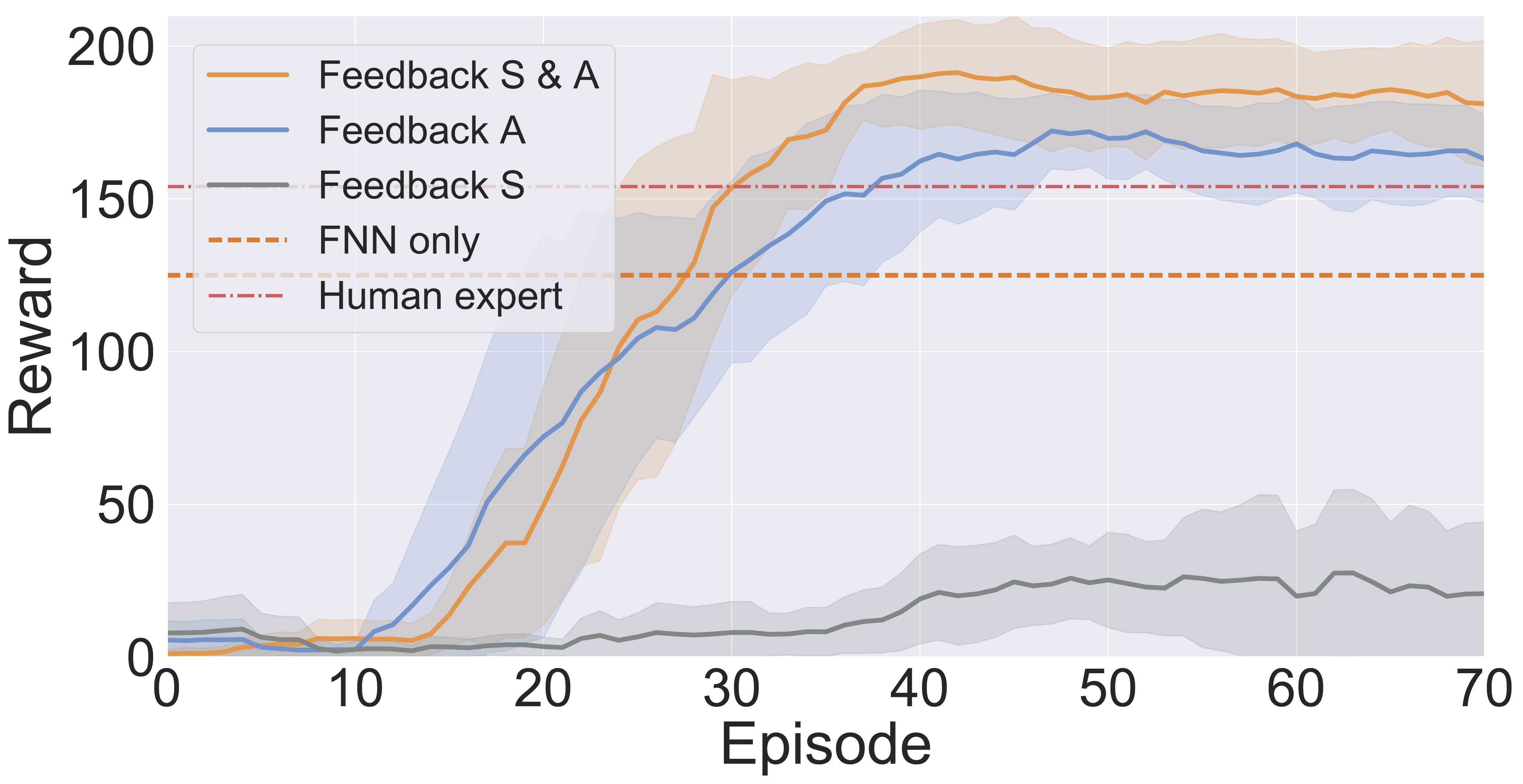}
	\caption{Effect of using different types of feedback in Bowling. The highest scores are achieved when both, feedback on states and actions are used. Providing feedback on states alone is not sufficient to guide the agent's learning process. The shaded region indicates the variance of the reward.}\label{SA}
\end{figure}

Figure \ref{SA} compares the effect of using different types of feedback for training the FNN.
We observe that feedback provided on both states and actions, or only on actions allows the agent to achieve super-human performance. 
Providing feedback on actions alone is only slightly worse in terms of average reward obtained and number of episodes required to perform better than a human expert. 
%While using feedback on both states and actions and using feedback on actions only can both surpass human expert, the performance of using feedback on action only is relatively worse in terms of both averaged reward and the episodes needed to outperform human expert.
In comparison, if only feedback on states is used to train the FNN, the agent is not able to obtain a high reward.
This indicates that feedback on states alone is not sufficient to efficiently guide the agent's learning process. A signal that incorporates feedback on actions taken at a state is better suited for this environment.
\begin{figure}
	\includegraphics[width=2.5 in]{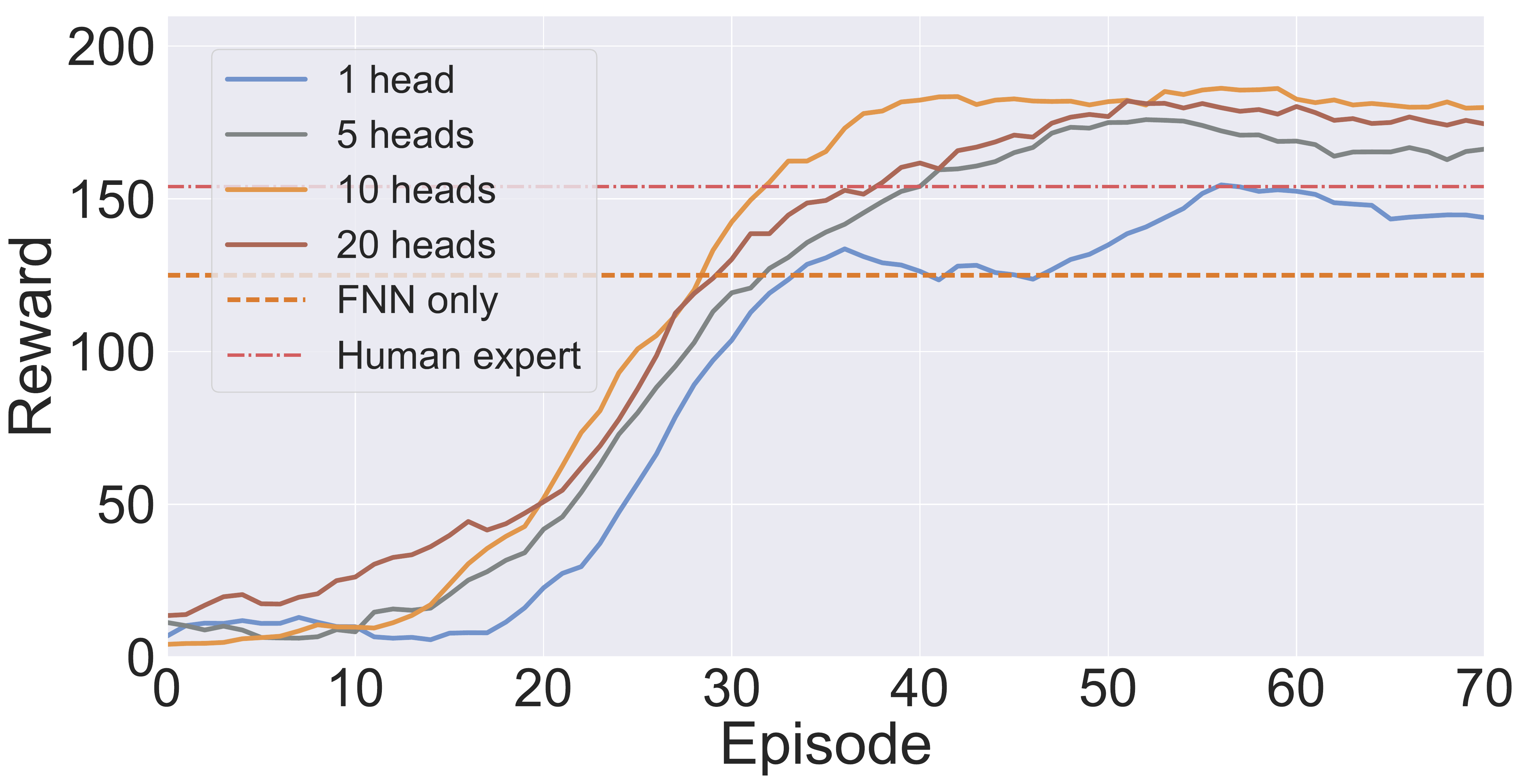}
	\caption{Effect of using different number of heads for the feedback neural network (FNN) in Bowling. When no ensemble is used ($1$ head), the performance of the FNN is worse than a human expert. Using $10$ heads for the FNN yields the best results. Too many heads may make the training process more difficult, and too few heads could mean that the FNN does not have the confidence to discard less useful outputs.}\label{N_head}
\end{figure}

\begin{figure}
	\includegraphics[width=2.0 in]{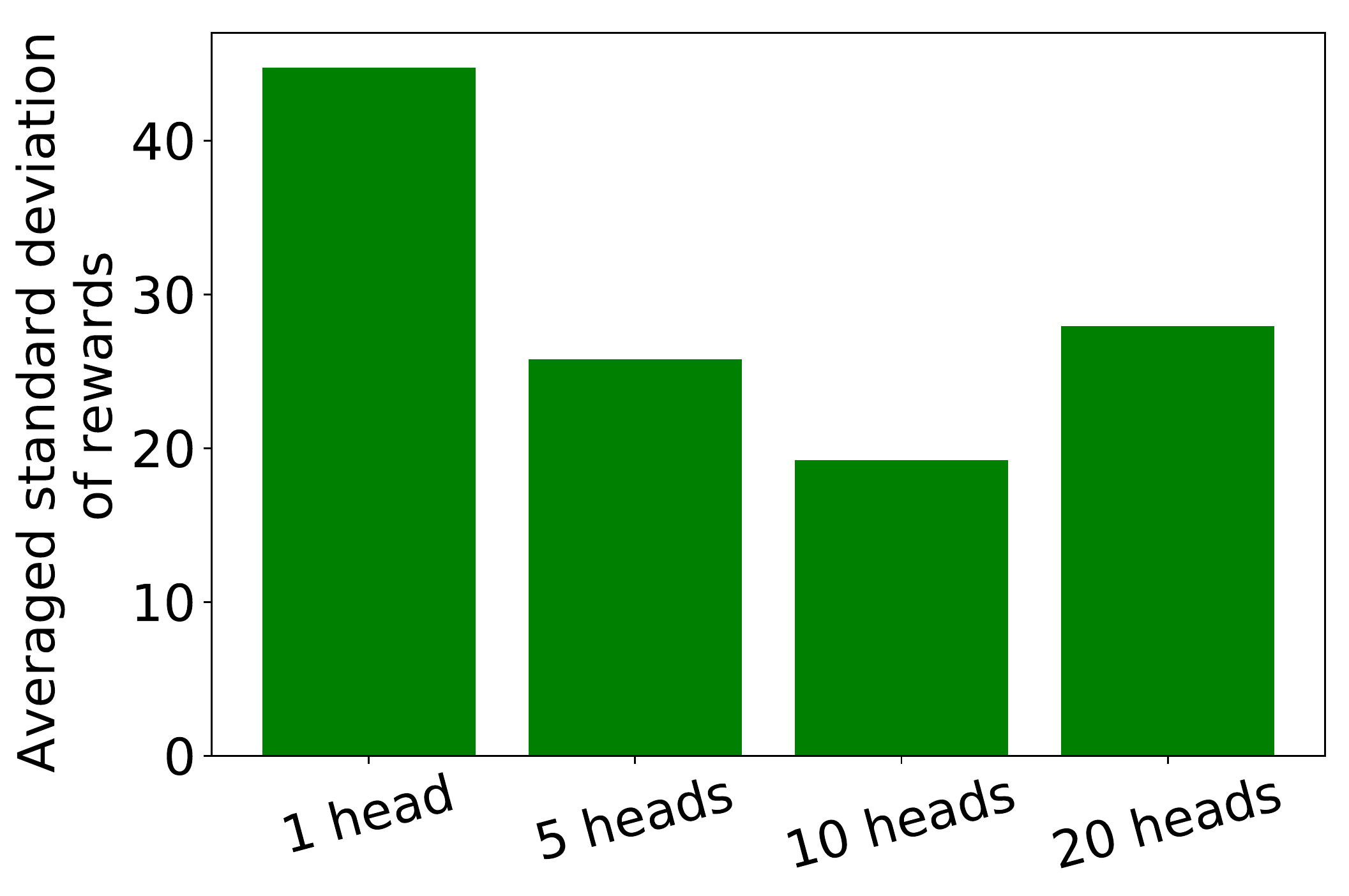}
	\caption{Averaged standard deviation of rewards in Bowling for different number of FNN heads. Using $10$ heads has the lowest average standard deviation. Having fewer heads could result in the FNN retaining less useful outputs, while having more heads makes training more difficult.}\label{N_head_std}
\end{figure}

In Figures \ref{N_head} and \ref{N_head_std}, we show the effect of varying the number of heads of the FNN on the learning process. 
When no ensemble is used, the average reward obtained is lower than that received by a human expert, and the (mean of the) standard deviation of the reward is high. 
This is an indication that in the absence of an ensemble, the agent could obtain a higher reward than a human expert player in some cases, but the large variance could also indicate that the performance is unstable. 
%Without emsemble, the averaged performance is lower than human expert and the mean std. of rewards is higher than emsemble method.
%This indicates that although the performance may be good at times without ensemble, it can be unstable (\textbf{sentence not clear}).
We observe that $K_a=K_s=10$ yields the best performance in terms of both variance (lowest) and averaged reward (highest).
Our conjecture is that having too many heads can make the training process more difficult while having too few heads might not be able to provide a good confidence value in order to discard some less useful outputs of the FNN.
\begin{figure}
	\includegraphics[width=2.5 in]{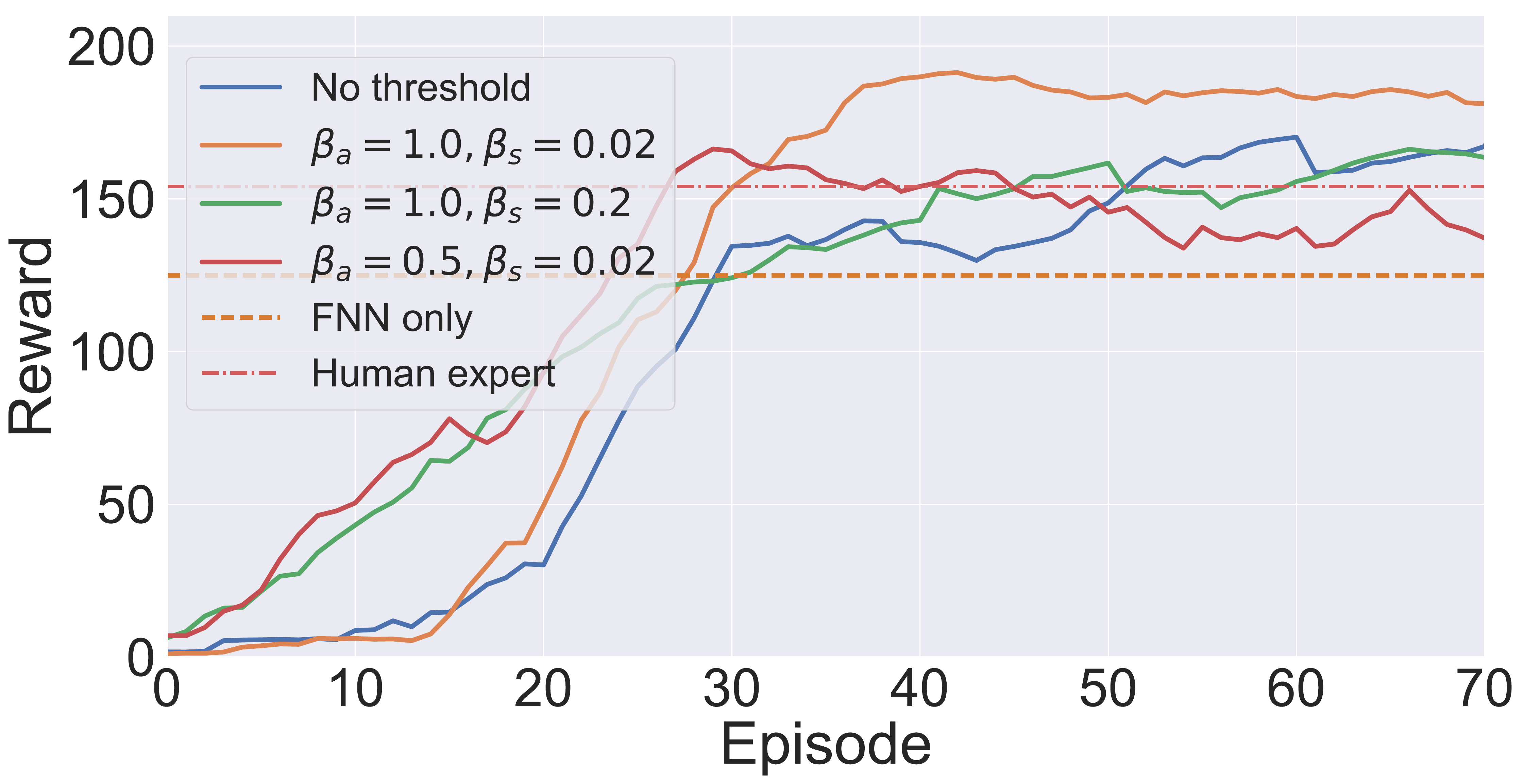}
	\caption{Effect of using different thresholds in Bowling. $\beta_a = 1.0$ and $\beta_s = 0.02$ results in the highest reward. $\beta_a = 1.0$ retains about $97\%$ of the FNN output, and $\beta_s = 0.02$ retains about $85\%$ of the FNN output.}\label{threshold}
\end{figure}

\begin{figure}
	\includegraphics[width=2.0 in]{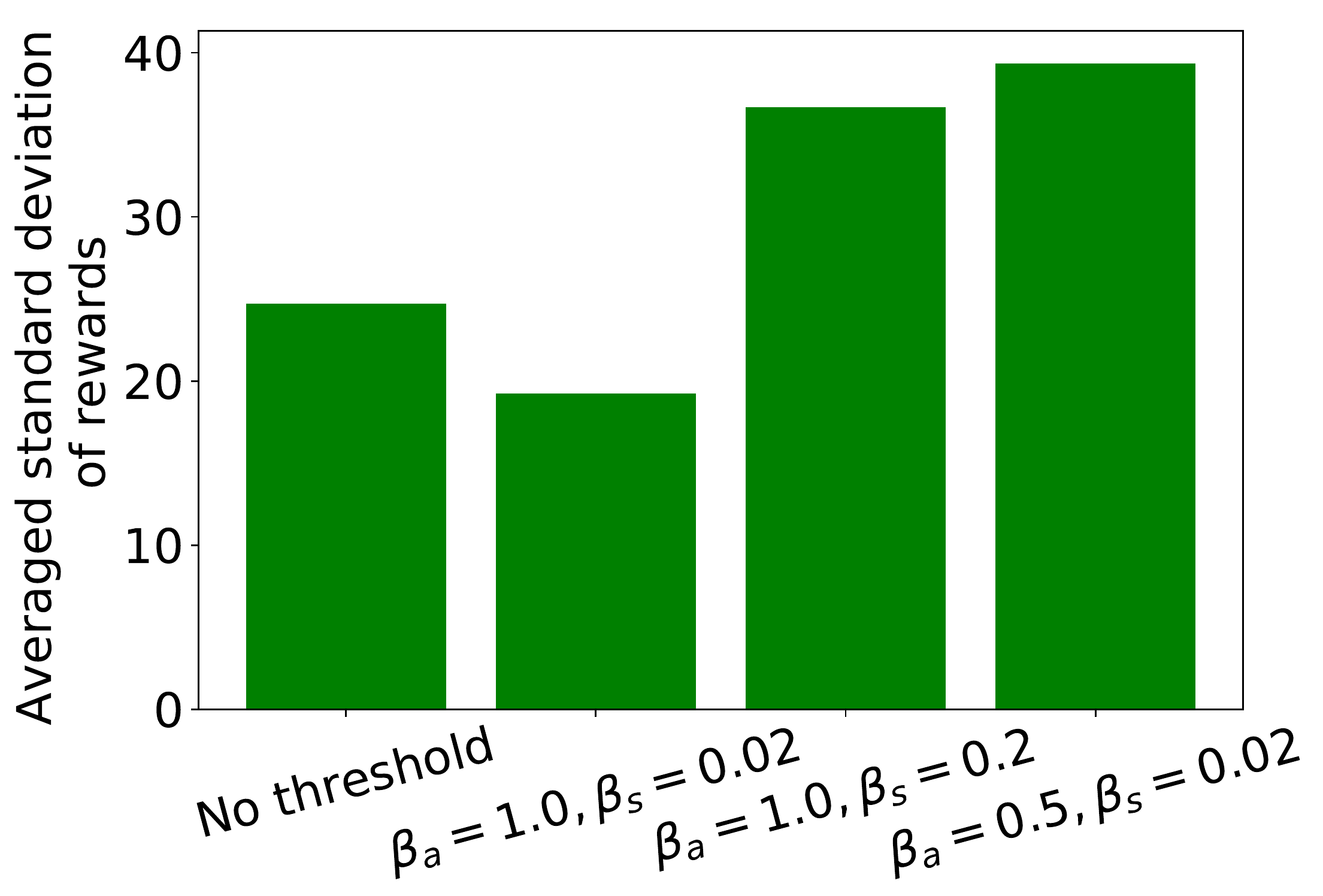}
	\caption{Averaged standard deviation of obtained rewards in Bowling for different threshold values. $\beta_a = 1.0$, $\beta_s = 0.02$ results in the lowest average standard deviation. Standard deviations for other threshold values is higher due to the FNN rejecting or accepting too many outputs of the FNN.}\label{threshold_std}
\end{figure}

Figures \ref{threshold} and \ref{threshold_std} show the effect of using different thresholds to eliminate some outputs of the FNN. 
\emph{FRESH} performs the best in terms of the average reward obtained, and variance of this reward when $(\beta_a=1.0,\beta_s=0.02)$. 
%With a proper threshold $(\beta_a=1.0,\beta_s=0.02)$, our algorithm performs the best in terms of the averaged reward.
In this case, setting $\beta_a=1.0$ retains about $97\%$ of the FNN outputs, and $\beta_s=0.02$ retains about $85\%$ of the FNN output. 
%In this case, setting $\beta_a=1.0$ roughly filters out $3\%$ of the output of FNN and setting $\beta_s=0.02$ roughly filters out $15\%$ of the output of FNN.
We observe that for some other threshold values, the performance is degraded due to rejecting or accepting too many outputs of the FNN (for example, $\beta_a=0.5$ rejects around $15\%$ and $\beta_s=0.2$ keeps around $97\%$ of the FNN outputs).
%We can also observe that setting an improper threshold ($\beta_a=0.5$ keeps around $\%85$) can be harmful due to rejecting or accepting too many ($\beta_s=0.2$ keeps around $97\%$) output of the FNN.

\section{Related Work} \label{Sec:RelatedWork}

The role of feedback provided by a human to an agent in RL settings has been studied by multiple researchers. 
A survey of recent research in using human guidance for deep RL tasks is presented in \cite{zhangleveraging}. 
We summarize related work in some of these techniques that are most relevant to this paper in this section. 

%\textbf{Reward shaping}: 
%One way of providing feedback is through \emph{reward shaping}, where 
Reward shaping modifies rewards supplied by the environment to accelerate the learning process of the agent  \cite{chang2006reinforcement, isbell2001social, thomaz2006reinforcement}. 
A framework called TAMER (Training an Agent Manually via Evaluative Reinforcement) that enabled \emph{shaping} (interactively training an agent via an external signal provided by a human) was presented in \cite{knox2009interactively}. 
This work was extended to enable human feedback to augment an RL agent that learned using an MDP reward signal in \cite{knox2010combining, knox2012reinforcement}. 
The outcome was that using this human feedback achieved a significant reduction in sample complexity. 
%However, there was not a formal rule that was used to incorporate this human feedback, and they used value-based methods in setups that involved only discrete actions. 
More recently, the authors of \cite{warnell2018deep} proposed Deep-TAMER, an architecture that extends the TAMER framework to environments with high-dimensional state spaces. 
In Deep-TAMER, the policy is given by a neural network that is trained via supervised learning using data from feedback provided by a human operator (while not considering the reward given by the environment). 
Another extension of TAMER, DQN-TAMER \cite{arakawa2018dqn}, modeled additional characteristics of human observers like inferring human reward from facial expressions. 
%The TAMER framework was not constrained by the ability of the human trainer to provide feedback, as long as the number of samples collected for learning an accurate representation of the system was sufficient. 
Signals provided by the human operator in TAMER was a numerical value indicative of how good the agent's behavior was, in the opinion of the operator. 
%This quantity is subjective, and may be very different from person-to-person. 
%\textbf{TAMER FEEDBACK is real-valued- may be VERY different from person-to-person}
%However, these papers do not provide estimates on the number of samples needed in order to train the RL agent to behave in a desired manner. 

Potential-based reward shaping (PBRS) methods used `potential functions' to accelerate the learning process, while preserving the identity of optimal policies \cite{Ng1999policy, Wiewiora2003principled, xiao2019potential}. 
The potential function was designed to encode `rules' of the environment of the RL agent. 
However, potential functions will typically need to be pre-specified.
%\textcolor{red}{, and moreover, it can be difficult to learn a potential function in high-dimensional spaces. 
This has restricted the use of PBRS to tabular/ low-dimensional state spaces \cite{Devlin2012express, snel2014learning}. 
%\textcolor{blue}{The cycling issue mentioned in section 4.1 can be resolved via PBRS, and hence transforming the output of FNN into potential rewards could be an interesting extension.}
The cycling problem (repeatedly visiting certain states) mention in Section 4.1 can be resolved by PBRS. 
This suggests that transforming the FNN output to a potential-based reward is an interesting direction of future research. 
%In some settings, it is possible to learn a potential function \cite{Devlin2012express, snel2014learning}, but this can be difficult to extend to high-dimensional state spaces. 
%However, these functions will typically need to be pre-specified, and they are difficult to learn. 

%\textbf{Preference-based learning}: 
Preference-based RL \cite{christiano2017deep} was used to communicate complex goals to allow systems to interact with real-world environments in a meaningful way. 
Although this approach required a human observer to compare trajectories and provide feedback during training, it alleviated the need for expert observers, since non-experts can easily compare and distinguish between `good' and `bad' trajectories. 
The human preferences were translated into a scalar reward, which was then used as a reward signal to an RL algorithm. 
This allowed the RL agent to directly learn from expert preferences. 
However, this approach is limited by assumptions on the existence of a (total) order among the set of trajectories.
%The paper by \cite{wirth2017survey} provides a survey of preference-based RL methods for the interested reader. 
A survey of preference-based RL methods is presented in \cite{wirth2017survey} for the interested reader.

%\textbf{Policy-dependent methods}: 
Consequences of the type and quality of human feedback that could result in forgetting of learned behavior was presented in COACH \cite{macglashan2017interactive}, which used policy-dependent human feedback to learn an optimal policy. 
The human feedback was interpreted as an unbiased estimate of the advantage function (i.e., the incremental value derived by an action in comparison to the current policy). 
The authors demonstrated that the role of human feedback is diminished as the agent learns the desired behavior. 
An extension to domains with large state spaces, Deep-COACH, was presented in \cite{arumugam2019deep}.
%
%A curiosity-based RL algorithm for sparse reward environments was presented in \cite{pathak2017curiosity}. 
%An intrinsic reward signal characterized the prediction error of the agent as a curiosity reward. 
%This method was also shown to be able to generalize to unseen scenarios. 
%A subsequent paper \cite{dubey2018investigating} provided a quantitative study of the roles of different types of human priors while playing video games. 
%One of their contributions was a taxonomy that ranked object priors in terms of their importance according to when they were acquired by a human. 
%In \cite{sukhbaatar2018intrinsic}, two versions of the same agent played with each other to learn about the environment in an unsupervised manner. 
%This was achieved by a reward structure that enabled the agents to automatically generate a curriculum of exploration. 

Another policy-based method to provide human feedback is \emph{policy shaping} \cite{griffith2013policy}. Here, feedback is a label on the optimality of an action, rather than a reward signal added to the reward from the environment. 
In this setup, the RL agent, in addition to receiving a reward from the environment, obtained an indication of whether the most recent action was correct or incorrect, and this label was used to infer the human's belief of the optimal policy in the current state. 
Further, it could be the case that there might be inconsistencies between the intended communication of the human and the information received by the agent, which results in the construction of a Bayes optimal feedback policy. 
Extensions to this work considered the effect of human attention \cite{faulkner2018policy}- which allows a robot (the RL agent) to learn desired behaviors faster by encouraging exploration of the environment in the presence of human supervision and exploiting actions for which positive feedback has been provided in states previously observed by the human supervisor- and settings where the robot can request feedback in states where it is not sure of the feedback received previously \cite{kessler2019active}. 
A similar approach that studied the interpretation of feedback strategies adopted by a human trainer as a probabilistic model was presented in \cite{loftin2016learning}, and this resulted in a strategy-aware Bayesian learning (SABL) algorithm. 

Demonstrations provided by a human operator were used to synthesize a `baseline policy' in Human-Agent Transfer (HAT) \cite{taylor2011integrating}. 
This baseline policy was then used to guide learning during the RL procedure. 
%Human-Agent Transfer (HAT) was introduced in \cite{taylor2011integrating}, where demonstrations provided by human are summarized into a baseline policy.
%In HAT, during reinforcement learning process, the baseline policy is fixed and used as a bias to guide learning.
CHAT \cite{wang2017improving} extended HAT to consider possible errors made while summarizing demonstrations, and used this uncertainty to improve performance. 
%The uncertainty in the source demonstrations is leveraged to improve performance.
Instead of providing demonstrations, the authors of \cite{ross2011reduction} presented DAGGER, an iterative imitation learning method which required a domain expert be available to provide correct actions during the entire learning process. 
A subsequent paper \cite{kelly2019hg} presented HG-DAGGER that predicted the performance of the agent using a threshold that modeled uncertainty.
%Under certain assumptions, DAGGER can be a no regret algorithm.
%One of the major challenges faced by RL algorithms that are used in high-dimensional state spaces is efficient exploration of its environment. 
%Most approaches that address the efficiency of exploration are designed for discrete-state environments with a small number of states \cite{jaksch2010near, guez2012efficient}. 
%For this reason, when RL is used in high-dimensional state spaces, exploration strategies used are often statistically inefficient \cite{mnih2015human}. 
%\textbf{Bootstrapped DQN}: 
%The authors of \cite{osband2016deep} proposed \emph{bootstrapped DQN}, which combined deep exploration with deep neural networks to accelerate the learning process. 
%The bootstrap principle is to approximate the distribution of a population by a sample distribution \cite{efron1994intro}. 
%This property allows us to use an ensemble of neural networks to effectively represent uncertainty in the model where each network in the ensemble produces a value indicative of its confidence in its output. 
%Another approach to represent uncertainty in deep learning models was presented in \cite{gal2016dropout}, where the authors established an equivalence between \emph{dropout} training in deep NNs and Bayesian inference in Gaussian processes. 
%However, in this paper, we prefer the bootstrapped DQN over dropout because of its lower time complexity during inference. 

%
%\textbf{related work on ensemble learning, ref. on feedback network, 1-2 sentences on dropout, }
%

\section{Conclusion} \label{Sec:Conclusion}

We presented \emph{FRESH}, a feedback-based reward shaping framework to effectively integrate human feedback with deep RL algorithms in high-dimensional state spaces. 
We used a feedback neural network to effectively generalize feedback signals provided by the human operator, and an ensemble of neural networks to represent the confidence of the neural network in its output. 
Our approach was evaluated on the \emph{Bowling} and \emph{Skiing} Atari games of the arcade learning environment. 
\emph{FRESH} performed much better than state-of-the-art deep learning algorithms in these environments. 
In \emph{Bowling}, \emph{FRESH} obtained an average score of $\mathbf{187}$, which was $\mathbf{21.4\%}$ higher than the score obtained by a human expert \cite{mnih2015human}. 
The highest score obtained by \emph{FRESH} in Bowling was $\mathbf{> 200}$. 
In \emph{Skiing}, \emph{FRESH} obtained an average score of $\mathbf{-4400}$, which was equal to that obtained by a human expert player.

\section*{Acknowledgement}
This work was supported by the U.S. Army Research Office and the Office of Naval Research via Grants W911NF-16-1-0485 and N00014-17-S-B001 respectively.

%%%%%%%%%%%%%%%%%%%%%%%%%%%%%%%%%%%%%%%%%%%%%%%%%%%%%%%%%%%%%%%%%%%%%%%%%%%%%%%%%%%%%%%%%%%%%%%%%%%%%%%%%
%% bibliography: see CFP for number of permitted pages

\bibliographystyle{ACM-Reference-Format}  % do not change this line!
\bibliography{AAMAS20_Bib}  % put name of your .bib file here


\begin{thebibliography}{50}

%%% the Web address.

\ifx \showCODEN    \undefined \def \showCODEN     #1{\unskip}     \fi
\ifx \showDOI      \undefined \def \showDOI       #1{#1}\fi
\ifx \showISBNx    \undefined \def \showISBNx     #1{\unskip}     \fi
\ifx \showISBNxiii \undefined \def \showISBNxiii  #1{\unskip}     \fi
\ifx \showISSN     \undefined \def \showISSN      #1{\unskip}     \fi
\ifx \showLCCN     \undefined \def \showLCCN      #1{\unskip}     \fi
\ifx \shownote     \undefined \def \shownote      #1{#1}          \fi
\ifx \showarticletitle \undefined \def \showarticletitle #1{#1}   \fi
\ifx \showURL      \undefined \def \showURL       {\relax}        \fi
% invisible to TeX
\providecommand\bibfield[2]{#2}
\providecommand\bibinfo[2]{#2}
\providecommand\natexlab[1]{#1}
\providecommand\showeprint[2][]{arXiv:#2}

\bibitem[\protect\citeauthoryear{Arakawa, Kobayashi, Unno, Tsuboi, and
  Maeda}{Arakawa et~al\mbox{.}}{2018}]%
        {arakawa2018dqn}
\bibfield{author}{\bibinfo{person}{Riku Arakawa}, \bibinfo{person}{Sosuke
  Kobayashi}, \bibinfo{person}{Yuya Unno}, \bibinfo{person}{Yuta Tsuboi}, {and}
  \bibinfo{person}{Shin-ichi Maeda}.} \bibinfo{year}{2018}\natexlab{}.
\newblock \showarticletitle{{DQN-TAMER}: {H}uman-in-the-Loop Reinforcement
  Learning with Intractable Feedback}.
\newblock \bibinfo{journal}{\emph{arXiv preprint arXiv:1810.11748}}
  (\bibinfo{year}{2018}).
\newblock


\bibitem[\protect\citeauthoryear{Arumugam, Lee, Saskin, and Littman}{Arumugam
  et~al\mbox{.}}{2019}]%
        {arumugam2019deep}
\bibfield{author}{\bibinfo{person}{Dilip Arumugam}, \bibinfo{person}{Jun~Ki
  Lee}, \bibinfo{person}{Sophie Saskin}, {and} \bibinfo{person}{Michael~L
  Littman}.} \bibinfo{year}{2019}\natexlab{}.
\newblock \showarticletitle{Deep reinforcement learning from policy-dependent
  human feedback}.
\newblock \bibinfo{journal}{\emph{arXiv preprint arXiv:1902.04257}}
  (\bibinfo{year}{2019}).
\newblock


\bibitem[\protect\citeauthoryear{{Bellemare}, {Naddaf}, {Veness}, and
  {Bowling}}{{Bellemare} et~al\mbox{.}}{2013}]%
        {bellemare13arcade}
\bibfield{author}{\bibinfo{person}{M.~G. {Bellemare}}, \bibinfo{person}{Y.
  {Naddaf}}, \bibinfo{person}{J. {Veness}}, {and} \bibinfo{person}{M.
  {Bowling}}.} \bibinfo{year}{2013}\natexlab{}.
\newblock \showarticletitle{The Arcade Learning Environment: An Evaluation
  Platform for General Agents}.
\newblock \bibinfo{journal}{\emph{Journal of Artificial Intelligence Research}}
   \bibinfo{volume}{47} (\bibinfo{year}{2013}), \bibinfo{pages}{253--279}.
\newblock


\bibitem[\protect\citeauthoryear{Berkenkamp, Turchetta, Schoellig, and
  Krause}{Berkenkamp et~al\mbox{.}}{2017}]%
        {berkenkamp2017safe}
\bibfield{author}{\bibinfo{person}{Felix Berkenkamp}, \bibinfo{person}{Matteo
  Turchetta}, \bibinfo{person}{Angela Schoellig}, {and}
  \bibinfo{person}{Andreas Krause}.} \bibinfo{year}{2017}\natexlab{}.
\newblock \showarticletitle{Safe model-based reinforcement learning with
  stability guarantees}. In \bibinfo{booktitle}{\emph{Advances in Neural
  Information Processing Systems}}. \bibinfo{pages}{908--918}.
\newblock


\bibitem[\protect\citeauthoryear{Chang}{Chang}{2006}]%
        {chang2006reinforcement}
\bibfield{author}{\bibinfo{person}{Hyeong~Soo Chang}.}
  \bibinfo{year}{2006}\natexlab{}.
\newblock \showarticletitle{Reinforcement learning with supervision by
  combining multiple learnings and expert advices}. In
  \bibinfo{booktitle}{\emph{Proceedings of the American Control Conference}}.
  \bibinfo{pages}{4159--4164}.
\newblock


\bibitem[\protect\citeauthoryear{Christiano, Leike, Brown, Martic, Legg, and
  Amodei}{Christiano et~al\mbox{.}}{2017}]%
        {christiano2017deep}
\bibfield{author}{\bibinfo{person}{Paul~F Christiano}, \bibinfo{person}{Jan
  Leike}, \bibinfo{person}{Tom Brown}, \bibinfo{person}{Miljan Martic},
  \bibinfo{person}{Shane Legg}, {and} \bibinfo{person}{Dario Amodei}.}
  \bibinfo{year}{2017}\natexlab{}.
\newblock \showarticletitle{Deep reinforcement learning from human
  preferences}. In \bibinfo{booktitle}{\emph{Advances in Neural Information
  Processing Systems}}. \bibinfo{pages}{4299--4307}.
\newblock


\bibitem[\protect\citeauthoryear{Efron and Tibshirani}{Efron and
  Tibshirani}{1994}]%
        {efron1994intro}
\bibfield{author}{\bibinfo{person}{Bradley Efron} {and} \bibinfo{person}{Robert
  Tibshirani}.} \bibinfo{year}{1994}\natexlab{}.
\newblock \bibinfo{booktitle}{\emph{An {I}ntroduction to the {B}ootstrap}}.
\newblock \bibinfo{publisher}{CRC Press}.
\newblock


\bibitem[\protect\citeauthoryear{Espeholt, Soyer, Munos, Simonyan, Mnih, Ward,
  Doron, Firoiu, Harley, Dunning, et~al\mbox{.}}{Espeholt
  et~al\mbox{.}}{2018}]%
        {espeholt2018impala}
\bibfield{author}{\bibinfo{person}{Lasse Espeholt}, \bibinfo{person}{Hubert
  Soyer}, \bibinfo{person}{Remi Munos}, \bibinfo{person}{Karen Simonyan},
  \bibinfo{person}{Volodymyr Mnih}, \bibinfo{person}{Tom Ward},
  \bibinfo{person}{Yotam Doron}, \bibinfo{person}{Vlad Firoiu},
  \bibinfo{person}{Tim Harley}, \bibinfo{person}{Iain Dunning},
  {et~al\mbox{.}}} \bibinfo{year}{2018}\natexlab{}.
\newblock \showarticletitle{IMPALA: Scalable Distributed Deep-RL with
  Importance Weighted Actor-Learner Architectures}. In
  \bibinfo{booktitle}{\emph{International Conference on Machine Learning}}.
  \bibinfo{pages}{1406--1415}.
\newblock


\bibitem[\protect\citeauthoryear{Faulkner, Short, and Thomaz}{Faulkner
  et~al\mbox{.}}{2018}]%
        {faulkner2018policy}
\bibfield{author}{\bibinfo{person}{Taylor~Kessler Faulkner},
  \bibinfo{person}{Elaine~Schaertl Short}, {and}
  \bibinfo{person}{Andrea~Lockerd Thomaz}.} \bibinfo{year}{2018}\natexlab{}.
\newblock \showarticletitle{Policy Shaping with Supervisory Attention Driven
  Exploration}. In \bibinfo{booktitle}{\emph{International Conference on
  Intelligent Robots and Systems}}. IEEE, \bibinfo{pages}{842--847}.
\newblock


\bibitem[\protect\citeauthoryear{Gal and Ghahramani}{Gal and
  Ghahramani}{2016}]%
        {gal2016dropout}
\bibfield{author}{\bibinfo{person}{Yarin Gal} {and} \bibinfo{person}{Zoubin
  Ghahramani}.} \bibinfo{year}{2016}\natexlab{}.
\newblock \showarticletitle{Dropout as {B}ayesian Approximation: {R}epresenting
  Model Uncertainty in Deep Learning}. In
  \bibinfo{booktitle}{\emph{International Conference on Machine Learning}}.
\newblock


\bibitem[\protect\citeauthoryear{Goodfellow, Bengio, and Courville}{Goodfellow
  et~al\mbox{.}}{2016}]%
        {goodfellow2016deep}
\bibfield{author}{\bibinfo{person}{Ian Goodfellow}, \bibinfo{person}{Yoshua
  Bengio}, {and} \bibinfo{person}{Aaron Courville}.}
  \bibinfo{year}{2016}\natexlab{}.
\newblock \bibinfo{booktitle}{\emph{Deep learning}}.
\newblock \bibinfo{publisher}{MIT press}.
\newblock


\bibitem[\protect\citeauthoryear{Griffith, Subramanian, Scholz, Isbell, and
  Thomaz}{Griffith et~al\mbox{.}}{2013}]%
        {griffith2013policy}
\bibfield{author}{\bibinfo{person}{Shane Griffith}, \bibinfo{person}{Kaushik
  Subramanian}, \bibinfo{person}{Jonathan Scholz}, \bibinfo{person}{Charles~L
  Isbell}, {and} \bibinfo{person}{Andrea~L Thomaz}.}
  \bibinfo{year}{2013}\natexlab{}.
\newblock \showarticletitle{Policy shaping: {I}ntegrating human feedback with
  reinforcement learning}. In \bibinfo{booktitle}{\emph{Advances in Neural
  Information Processing Systems}}. \bibinfo{pages}{2625--2633}.
\newblock


\bibitem[\protect\citeauthoryear{Gu, Lillicrap, Sutskever, and Levine}{Gu
  et~al\mbox{.}}{2016}]%
        {gu2016continuous}
\bibfield{author}{\bibinfo{person}{Shixiang Gu}, \bibinfo{person}{Timothy
  Lillicrap}, \bibinfo{person}{Ilya Sutskever}, {and} \bibinfo{person}{Sergey
  Levine}.} \bibinfo{year}{2016}\natexlab{}.
\newblock \showarticletitle{Continuous deep {Q}-learning with model-based
  acceleration}. In \bibinfo{booktitle}{\emph{International Conference on
  Machine Learning}}. \bibinfo{pages}{2829--2838}.
\newblock


\bibitem[\protect\citeauthoryear{Hafner and Riedmiller}{Hafner and
  Riedmiller}{2011}]%
        {hafner2011reinforcement}
\bibfield{author}{\bibinfo{person}{Roland Hafner} {and} \bibinfo{person}{Martin
  Riedmiller}.} \bibinfo{year}{2011}\natexlab{}.
\newblock \showarticletitle{Reinforcement learning in feedback control}.
\newblock \bibinfo{journal}{\emph{Machine Learning}}  \bibinfo{volume}{84}
  (\bibinfo{year}{2011}), \bibinfo{pages}{137--169}.
\newblock


\bibitem[\protect\citeauthoryear{Harutyunyan, Devlin, Vrancx, and
  Now{\'e}}{Harutyunyan et~al\mbox{.}}{2015}]%
        {Devlin2012express}
\bibfield{author}{\bibinfo{person}{Anna Harutyunyan}, \bibinfo{person}{Sam
  Devlin}, \bibinfo{person}{Peter Vrancx}, {and} \bibinfo{person}{Ann
  Now{\'e}}.} \bibinfo{year}{2015}\natexlab{}.
\newblock \showarticletitle{Expressing Arbitrary Reward Functions as
  Potential-Based Advice}. In \bibinfo{booktitle}{\emph{AAAI}}.
  \bibinfo{pages}{2652--2658}.
\newblock


\bibitem[\protect\citeauthoryear{Hessel, Modayil, Van~Hasselt, Schaul,
  Ostrovski, Dabney, Horgan, Piot, Azar, and Silver}{Hessel
  et~al\mbox{.}}{2018}]%
        {hessel2018rainbow}
\bibfield{author}{\bibinfo{person}{Matteo Hessel}, \bibinfo{person}{Joseph
  Modayil}, \bibinfo{person}{Hado Van~Hasselt}, \bibinfo{person}{Tom Schaul},
  \bibinfo{person}{Georg Ostrovski}, \bibinfo{person}{Will Dabney},
  \bibinfo{person}{Dan Horgan}, \bibinfo{person}{Bilal Piot},
  \bibinfo{person}{Mohammad Azar}, {and} \bibinfo{person}{David Silver}.}
  \bibinfo{year}{2018}\natexlab{}.
\newblock \showarticletitle{Rainbow: Combining improvements in deep
  reinforcement learning}. In \bibinfo{booktitle}{\emph{AAAI}}.
\newblock


\bibitem[\protect\citeauthoryear{Isbell, Shelton, Kearns, Singh, and
  Stone}{Isbell et~al\mbox{.}}{2001}]%
        {isbell2001social}
\bibfield{author}{\bibinfo{person}{Charles Isbell},
  \bibinfo{person}{Christian~R Shelton}, \bibinfo{person}{Michael Kearns},
  \bibinfo{person}{Satinder Singh}, {and} \bibinfo{person}{Peter Stone}.}
  \bibinfo{year}{2001}\natexlab{}.
\newblock \showarticletitle{A social reinforcement learning agent}. In
  \bibinfo{booktitle}{\emph{Proceedings of the International Conference on
  Autonomous Agents}}. \bibinfo{pages}{377--384}.
\newblock


\bibitem[\protect\citeauthoryear{Kamalapurkar, Walters, and Dixon}{Kamalapurkar
  et~al\mbox{.}}{2016}]%
        {kamalapurkar2016model}
\bibfield{author}{\bibinfo{person}{Rushikesh Kamalapurkar},
  \bibinfo{person}{Patrick Walters}, {and} \bibinfo{person}{Warren~E Dixon}.}
  \bibinfo{year}{2016}\natexlab{}.
\newblock \showarticletitle{Model-based reinforcement learning for approximate
  optimal regulation}.
\newblock \bibinfo{journal}{\emph{Automatica}}  \bibinfo{volume}{64}
  (\bibinfo{year}{2016}), \bibinfo{pages}{94--104}.
\newblock


\bibitem[\protect\citeauthoryear{Kelly, Sidrane, Driggs-Campbell, and
  Kochenderfer}{Kelly et~al\mbox{.}}{2019}]%
        {kelly2019hg}
\bibfield{author}{\bibinfo{person}{Michael Kelly}, \bibinfo{person}{Chelsea
  Sidrane}, \bibinfo{person}{Katherine Driggs-Campbell}, {and}
  \bibinfo{person}{Mykel~J Kochenderfer}.} \bibinfo{year}{2019}\natexlab{}.
\newblock \showarticletitle{HG-DAgger: Interactive Imitation Learning with
  Human Experts}. In \bibinfo{booktitle}{\emph{International Conference on
  Robotics and Automation}}. IEEE, \bibinfo{pages}{8077--8083}.
\newblock


\bibitem[\protect\citeauthoryear{Kessler~Faulkner, Gutierrez, Short, Hoffman,
  and Thomaz}{Kessler~Faulkner et~al\mbox{.}}{2019}]%
        {kessler2019active}
\bibfield{author}{\bibinfo{person}{Taylor Kessler~Faulkner},
  \bibinfo{person}{Reymundo~A Gutierrez}, \bibinfo{person}{Elaine~Schaertl
  Short}, \bibinfo{person}{Guy Hoffman}, {and} \bibinfo{person}{Andrea~L
  Thomaz}.} \bibinfo{year}{2019}\natexlab{}.
\newblock \showarticletitle{Active Attention-Modified Policy Shaping:
  {S}ocially Interactive Agents Track}. In \bibinfo{booktitle}{\emph{Autonomous
  Agents and MultiAgent Systems}}. \bibinfo{pages}{728--736}.
\newblock


\bibitem[\protect\citeauthoryear{Knox and Stone}{Knox and Stone}{2009}]%
        {knox2009interactively}
\bibfield{author}{\bibinfo{person}{W~Bradley Knox} {and} \bibinfo{person}{Peter
  Stone}.} \bibinfo{year}{2009}\natexlab{}.
\newblock \showarticletitle{Interactively shaping agents via human
  reinforcement: The {TAMER} framework}. In
  \bibinfo{booktitle}{\emph{International Conference on Knowledge Capture}}.
  \bibinfo{pages}{9--16}.
\newblock


\bibitem[\protect\citeauthoryear{Knox and Stone}{Knox and Stone}{2010}]%
        {knox2010combining}
\bibfield{author}{\bibinfo{person}{W~Bradley Knox} {and} \bibinfo{person}{Peter
  Stone}.} \bibinfo{year}{2010}\natexlab{}.
\newblock \showarticletitle{Combining manual feedback with subsequent {MDP}
  reward signals for reinforcement learning}. In
  \bibinfo{booktitle}{\emph{Autonomous Agents and Multiagent Systems}}.
  \bibinfo{pages}{5--12}.
\newblock


\bibitem[\protect\citeauthoryear{Knox and Stone}{Knox and Stone}{2012}]%
        {knox2012reinforcement}
\bibfield{author}{\bibinfo{person}{W~Bradley Knox} {and} \bibinfo{person}{Peter
  Stone}.} \bibinfo{year}{2012}\natexlab{}.
\newblock \showarticletitle{Reinforcement learning from simultaneous human and
  {MDP} reward}. In \bibinfo{booktitle}{\emph{Autonomous Agents and Multiagent
  Systems}}. \bibinfo{pages}{475--482}.
\newblock


\bibitem[\protect\citeauthoryear{Lakshminarayanan, Pritzel, and
  Blundell}{Lakshminarayanan et~al\mbox{.}}{2017}]%
        {lakshminarayanan2017simple}
\bibfield{author}{\bibinfo{person}{Balaji Lakshminarayanan},
  \bibinfo{person}{Alexander Pritzel}, {and} \bibinfo{person}{Charles
  Blundell}.} \bibinfo{year}{2017}\natexlab{}.
\newblock \showarticletitle{Simple and scalable predictive uncertainty
  estimation using deep ensembles}. In \bibinfo{booktitle}{\emph{Advances in
  Neural Information Processing Systems}}. \bibinfo{pages}{6402--6413}.
\newblock


\bibitem[\protect\citeauthoryear{Levine, Finn, Darrell, and Abbeel}{Levine
  et~al\mbox{.}}{2016}]%
        {levine2016end}
\bibfield{author}{\bibinfo{person}{Sergey Levine}, \bibinfo{person}{Chelsea
  Finn}, \bibinfo{person}{Trevor Darrell}, {and} \bibinfo{person}{Pieter
  Abbeel}.} \bibinfo{year}{2016}\natexlab{}.
\newblock \showarticletitle{End-to-end training of deep visuomotor policies}.
\newblock \bibinfo{journal}{\emph{The Journal of Machine Learning Research}}
  \bibinfo{volume}{17}, \bibinfo{number}{1} (\bibinfo{year}{2016}),
  \bibinfo{pages}{1334--1373}.
\newblock


\bibitem[\protect\citeauthoryear{Lillicrap, Hunt, Pritzel, Heess, Erez, Tassa,
  Silver, and Wierstra}{Lillicrap et~al\mbox{.}}{2016}]%
        {lillicrap2016continuous}
\bibfield{author}{\bibinfo{person}{Timothy~P Lillicrap},
  \bibinfo{person}{Jonathan~J Hunt}, \bibinfo{person}{Alexander Pritzel},
  \bibinfo{person}{Nicolas Heess}, \bibinfo{person}{Tom Erez},
  \bibinfo{person}{Yuval Tassa}, \bibinfo{person}{David Silver}, {and}
  \bibinfo{person}{Daan Wierstra}.} \bibinfo{year}{2016}\natexlab{}.
\newblock \showarticletitle{Continuous control with deep reinforcement
  learning}. In \bibinfo{booktitle}{\emph{International Conference on Learning
  and Representations}}.
\newblock


\bibitem[\protect\citeauthoryear{Loftin, Peng, MacGlashan, Littman, Taylor,
  Huang, and Roberts}{Loftin et~al\mbox{.}}{2016}]%
        {loftin2016learning}
\bibfield{author}{\bibinfo{person}{Robert Loftin}, \bibinfo{person}{Bei Peng},
  \bibinfo{person}{James MacGlashan}, \bibinfo{person}{Michael~L Littman},
  \bibinfo{person}{Matthew~E Taylor}, \bibinfo{person}{Jeff Huang}, {and}
  \bibinfo{person}{David~L Roberts}.} \bibinfo{year}{2016}\natexlab{}.
\newblock \showarticletitle{Learning behaviors via human-delivered discrete
  feedback: {M}odeling implicit feedback strategies to speed up learning}.
\newblock \bibinfo{journal}{\emph{Autonomous agents and multi-agent systems}}
  \bibinfo{volume}{30}, \bibinfo{number}{1} (\bibinfo{year}{2016}),
  \bibinfo{pages}{30--59}.
\newblock


\bibitem[\protect\citeauthoryear{MacGlashan, Ho, Loftin, Peng, Wang, Roberts,
  Taylor, and Littman}{MacGlashan et~al\mbox{.}}{2017}]%
        {macglashan2017interactive}
\bibfield{author}{\bibinfo{person}{James MacGlashan}, \bibinfo{person}{Mark~K
  Ho}, \bibinfo{person}{Robert Loftin}, \bibinfo{person}{Bei Peng},
  \bibinfo{person}{Guan Wang}, \bibinfo{person}{David~L Roberts},
  \bibinfo{person}{Matthew~E Taylor}, {and} \bibinfo{person}{Michael~L
  Littman}.} \bibinfo{year}{2017}\natexlab{}.
\newblock \showarticletitle{Interactive learning from policy-dependent human
  feedback}. In \bibinfo{booktitle}{\emph{International Conference on Machine
  Learning}}. \bibinfo{pages}{2285--2294}.
\newblock


\bibitem[\protect\citeauthoryear{Mnih, Kavukcuoglu, Silver, Rusu, Veness,
  Bellemare, Graves, Riedmiller, Fidjeland, Ostrovski, et~al\mbox{.}}{Mnih
  et~al\mbox{.}}{2015}]%
        {mnih2015human}
\bibfield{author}{\bibinfo{person}{Volodymyr Mnih}, \bibinfo{person}{Koray
  Kavukcuoglu}, \bibinfo{person}{David Silver}, \bibinfo{person}{Andrei~A
  Rusu}, \bibinfo{person}{Joel Veness}, \bibinfo{person}{Marc~G Bellemare},
  \bibinfo{person}{Alex Graves}, \bibinfo{person}{Martin Riedmiller},
  \bibinfo{person}{Andreas~K Fidjeland}, \bibinfo{person}{Georg Ostrovski},
  {et~al\mbox{.}}} \bibinfo{year}{2015}\natexlab{}.
\newblock \showarticletitle{Human-level control through deep reinforcement
  learning}.
\newblock \bibinfo{journal}{\emph{Nature}} \bibinfo{volume}{518},
  \bibinfo{number}{7540} (\bibinfo{year}{2015}), \bibinfo{pages}{529}.
\newblock


\bibitem[\protect\citeauthoryear{Neal}{Neal}{2012}]%
        {neal2012bayesian}
\bibfield{author}{\bibinfo{person}{Radford~M Neal}.}
  \bibinfo{year}{2012}\natexlab{}.
\newblock \bibinfo{booktitle}{\emph{Bayesian learning for neural networks}}.
  Vol.~\bibinfo{volume}{118}.
\newblock \bibinfo{publisher}{Springer Science \& Business Media}.
\newblock


\bibitem[\protect\citeauthoryear{Ng, Harada, and Russell}{Ng
  et~al\mbox{.}}{1999}]%
        {Ng1999policy}
\bibfield{author}{\bibinfo{person}{Andrew~Y Ng}, \bibinfo{person}{Daishi
  Harada}, {and} \bibinfo{person}{Stuart Russell}.}
  \bibinfo{year}{1999}\natexlab{}.
\newblock \showarticletitle{Policy invariance under reward transformations:
  {T}heory and application to reward shaping}. In
  \bibinfo{booktitle}{\emph{International Conference on Machine Learning}}.
\newblock


\bibitem[\protect\citeauthoryear{Osband, Blundell, Pritzel, and Roy}{Osband
  et~al\mbox{.}}{2016}]%
        {osband2016deep}
\bibfield{author}{\bibinfo{person}{Ian Osband}, \bibinfo{person}{Charles
  Blundell}, \bibinfo{person}{Alexander Pritzel}, {and}
  \bibinfo{person}{Benjamin~Van Roy}.} \bibinfo{year}{2016}\natexlab{}.
\newblock \showarticletitle{Deep exploration via bootstrapped {DQN}}. In
  \bibinfo{booktitle}{\emph{Advances in Neural Information Processing
  Systems}}.
\newblock


\bibitem[\protect\citeauthoryear{Pathak, Agrawal, Efros, and Darrell}{Pathak
  et~al\mbox{.}}{2017}]%
        {pathak2017curiosity}
\bibfield{author}{\bibinfo{person}{Deepak Pathak}, \bibinfo{person}{Pulkit
  Agrawal}, \bibinfo{person}{Alexei~A Efros}, {and} \bibinfo{person}{Trevor
  Darrell}.} \bibinfo{year}{2017}\natexlab{}.
\newblock \showarticletitle{Curiosity-driven exploration by self-supervised
  prediction}. In \bibinfo{booktitle}{\emph{International Conference on Machine
  Learning}}.
\newblock


\bibitem[\protect\citeauthoryear{Puterman}{Puterman}{2014}]%
        {puterman2014markov}
\bibfield{author}{\bibinfo{person}{Martin~L Puterman}.}
  \bibinfo{year}{2014}\natexlab{}.
\newblock \bibinfo{booktitle}{\emph{Markov Decision Processes: {D}iscrete
  Stochastic Dynamic Programming}}.
\newblock \bibinfo{publisher}{John Wiley \& Sons}.
\newblock


\bibitem[\protect\citeauthoryear{Ross, Gordon, and Bagnell}{Ross
  et~al\mbox{.}}{2011}]%
        {ross2011reduction}
\bibfield{author}{\bibinfo{person}{St{\'e}phane Ross},
  \bibinfo{person}{Geoffrey Gordon}, {and} \bibinfo{person}{Drew Bagnell}.}
  \bibinfo{year}{2011}\natexlab{}.
\newblock \showarticletitle{A reduction of imitation learning and structured
  prediction to no-regret online learning}. In
  \bibinfo{booktitle}{\emph{Proceedings of the fourteenth international
  conference on artificial intelligence and statistics}}.
  \bibinfo{pages}{627--635}.
\newblock


\bibitem[\protect\citeauthoryear{Schulman, Levine, Abbeel, Jordan, and
  Moritz}{Schulman et~al\mbox{.}}{2015}]%
        {schulman2015trust}
\bibfield{author}{\bibinfo{person}{John Schulman}, \bibinfo{person}{Sergey
  Levine}, \bibinfo{person}{Pieter Abbeel}, \bibinfo{person}{Michael Jordan},
  {and} \bibinfo{person}{Philipp Moritz}.} \bibinfo{year}{2015}\natexlab{}.
\newblock \showarticletitle{Trust region policy optimization}. In
  \bibinfo{booktitle}{\emph{International Conference on Machine Learning}}.
  \bibinfo{pages}{1889--1897}.
\newblock


\bibitem[\protect\citeauthoryear{Silver, Huang, Maddison, Guez, Sifre, Van
  Den~Driessche, Schrittwieser, Antonoglou, Panneershelvam, Lanctot, Dieleman,
  Grewe, Nham, Kalchbrenner, Sutskever, Lillicrap, Leach, Kavukcuoglu, Graepel,
  and Hassabis}{Silver et~al\mbox{.}}{2016}]%
        {silver2016mastering}
\bibfield{author}{\bibinfo{person}{David Silver}, \bibinfo{person}{Aja Huang},
  \bibinfo{person}{Chris~J Maddison}, \bibinfo{person}{Arthur Guez},
  \bibinfo{person}{Laurent Sifre}, \bibinfo{person}{George Van Den~Driessche},
  \bibinfo{person}{Julian Schrittwieser}, \bibinfo{person}{Ioannis Antonoglou},
  \bibinfo{person}{Veda Panneershelvam}, \bibinfo{person}{Marc Lanctot},
  \bibinfo{person}{Sander Dieleman}, \bibinfo{person}{Dominik Grewe},
  \bibinfo{person}{John Nham}, \bibinfo{person}{Nal Kalchbrenner},
  \bibinfo{person}{Ilya Sutskever}, \bibinfo{person}{Timothy Lillicrap},
  \bibinfo{person}{Madeleine Leach}, \bibinfo{person}{Koray Kavukcuoglu},
  \bibinfo{person}{Thore Graepel}, {and} \bibinfo{person}{Demis Hassabis}.}
  \bibinfo{year}{2016}\natexlab{}.
\newblock \showarticletitle{Mastering the game of {G}o with deep neural
  networks and tree search}.
\newblock \bibinfo{journal}{\emph{Nature}} \bibinfo{volume}{529},
  \bibinfo{number}{7587} (\bibinfo{year}{2016}).
\newblock


\bibitem[\protect\citeauthoryear{Snel and Whiteson}{Snel and Whiteson}{2014}]%
        {snel2014learning}
\bibfield{author}{\bibinfo{person}{Matthijs Snel} {and} \bibinfo{person}{Shimon
  Whiteson}.} \bibinfo{year}{2014}\natexlab{}.
\newblock \showarticletitle{Learning potential functions and their
  representations for multi-task reinforcement learning}.
\newblock \bibinfo{journal}{\emph{Autonomous Agents and Multi-Agent Systems}}
  \bibinfo{volume}{28}, \bibinfo{number}{4} (\bibinfo{year}{2014}),
  \bibinfo{pages}{637--681}.
\newblock


\bibitem[\protect\citeauthoryear{Sutton and Barto}{Sutton and Barto}{2018}]%
        {sutton2018reinforcement}
\bibfield{author}{\bibinfo{person}{Richard~S Sutton} {and}
  \bibinfo{person}{Andrew~G Barto}.} \bibinfo{year}{2018}\natexlab{}.
\newblock \bibinfo{booktitle}{\emph{Reinforcement Learning: {A}n
  Introduction}}.
\newblock \bibinfo{publisher}{MIT Press}.
\newblock


\bibitem[\protect\citeauthoryear{Taylor, Suay, and Chernova}{Taylor
  et~al\mbox{.}}{2011}]%
        {taylor2011integrating}
\bibfield{author}{\bibinfo{person}{Matthew~E Taylor},
  \bibinfo{person}{Halit~Bener Suay}, {and} \bibinfo{person}{Sonia Chernova}.}
  \bibinfo{year}{2011}\natexlab{}.
\newblock \showarticletitle{Integrating reinforcement learning with human
  demonstrations of varying ability}. In \bibinfo{booktitle}{\emph{The 10th
  International Conference on Autonomous Agents and Multiagent Systems-Volume
  2}}. International Foundation for Autonomous Agents and Multiagent Systems,
  \bibinfo{pages}{617--624}.
\newblock


\bibitem[\protect\citeauthoryear{Thomaz and Breazeal}{Thomaz and
  Breazeal}{2006}]%
        {thomaz2006reinforcement}
\bibfield{author}{\bibinfo{person}{Andrea~Lockerd Thomaz} {and}
  \bibinfo{person}{Cynthia Breazeal}.} \bibinfo{year}{2006}\natexlab{}.
\newblock \showarticletitle{Reinforcement learning with human teachers:
  {E}vidence of feedback and guidance with implications for learning
  performance}. In \bibinfo{booktitle}{\emph{AAAI}}.
  \bibinfo{pages}{1000--1005}.
\newblock


\bibitem[\protect\citeauthoryear{Van~Hasselt, Guez, and Silver}{Van~Hasselt
  et~al\mbox{.}}{2016}]%
        {van2016deep}
\bibfield{author}{\bibinfo{person}{Hado Van~Hasselt}, \bibinfo{person}{Arthur
  Guez}, {and} \bibinfo{person}{David Silver}.}
  \bibinfo{year}{2016}\natexlab{}.
\newblock \showarticletitle{Deep reinforcement learning with double
  {Q}-learning}. In \bibinfo{booktitle}{\emph{AAAI}}.
\newblock


\bibitem[\protect\citeauthoryear{Wang, Schaul, Hessel, Hasselt, Lanctot, and
  Freitas}{Wang et~al\mbox{.}}{2016}]%
        {wang2016dueling}
\bibfield{author}{\bibinfo{person}{Ziyu Wang}, \bibinfo{person}{Tom Schaul},
  \bibinfo{person}{Matteo Hessel}, \bibinfo{person}{Hado Hasselt},
  \bibinfo{person}{Marc Lanctot}, {and} \bibinfo{person}{Nando Freitas}.}
  \bibinfo{year}{2016}\natexlab{}.
\newblock \showarticletitle{Dueling Network Architectures for Deep
  Reinforcement Learning}. In \bibinfo{booktitle}{\emph{International
  Conference on Machine Learning}}. \bibinfo{pages}{1995--2003}.
\newblock


\bibitem[\protect\citeauthoryear{Wang and Taylor}{Wang and Taylor}{2017}]%
        {wang2017improving}
\bibfield{author}{\bibinfo{person}{Zhaodong Wang} {and}
  \bibinfo{person}{Matthew~E Taylor}.} \bibinfo{year}{2017}\natexlab{}.
\newblock \showarticletitle{Improving Reinforcement Learning with
  Confidence-Based Demonstrations.}. In \bibinfo{booktitle}{\emph{IJCAI}}.
  \bibinfo{pages}{3027--3033}.
\newblock


\bibitem[\protect\citeauthoryear{Warnell, Waytowich, Lawhern, and
  Stone}{Warnell et~al\mbox{.}}{2018}]%
        {warnell2018deep}
\bibfield{author}{\bibinfo{person}{Garrett Warnell}, \bibinfo{person}{Nicholas
  Waytowich}, \bibinfo{person}{Vernon Lawhern}, {and} \bibinfo{person}{Peter
  Stone}.} \bibinfo{year}{2018}\natexlab{}.
\newblock \showarticletitle{Deep {TAMER}: {I}nteractive agent shaping in
  high-dimensional state spaces}. In \bibinfo{booktitle}{\emph{AAAI}}.
\newblock


\bibitem[\protect\citeauthoryear{Watkins and Dayan}{Watkins and Dayan}{1992}]%
        {watkins1992q}
\bibfield{author}{\bibinfo{person}{Christopher~JCH Watkins} {and}
  \bibinfo{person}{Peter Dayan}.} \bibinfo{year}{1992}\natexlab{}.
\newblock \showarticletitle{Q-learning}.
\newblock \bibinfo{journal}{\emph{Machine learning}} \bibinfo{volume}{8},
  \bibinfo{number}{3-4} (\bibinfo{year}{1992}), \bibinfo{pages}{279--292}.
\newblock


\bibitem[\protect\citeauthoryear{Wiewiora, Cottrell, and Elkan}{Wiewiora
  et~al\mbox{.}}{2003}]%
        {Wiewiora2003principled}
\bibfield{author}{\bibinfo{person}{Eric Wiewiora}, \bibinfo{person}{Garrison~W
  Cottrell}, {and} \bibinfo{person}{Charles Elkan}.}
  \bibinfo{year}{2003}\natexlab{}.
\newblock \showarticletitle{Principled methods for advising reinforcement
  learning agents}. In \bibinfo{booktitle}{\emph{International Conference on
  Machine Learning}}. \bibinfo{pages}{792--799}.
\newblock


\bibitem[\protect\citeauthoryear{Wirth, Akrour, Neumann, and
  F{\"u}rnkranz}{Wirth et~al\mbox{.}}{2017}]%
        {wirth2017survey}
\bibfield{author}{\bibinfo{person}{Christian Wirth}, \bibinfo{person}{Riad
  Akrour}, \bibinfo{person}{Gerhard Neumann}, {and} \bibinfo{person}{Johannes
  F{\"u}rnkranz}.} \bibinfo{year}{2017}\natexlab{}.
\newblock \showarticletitle{A survey of preference-based reinforcement learning
  methods}.
\newblock \bibinfo{journal}{\emph{The Journal of Machine Learning Research}}
  \bibinfo{volume}{18}, \bibinfo{number}{1} (\bibinfo{year}{2017}),
  \bibinfo{pages}{4945--4990}.
\newblock


\bibitem[\protect\citeauthoryear{Xiao, Ramasubramanian, Clark, Hajishirzi,
  Bushnell, and Poovendran}{Xiao et~al\mbox{.}}{2019}]%
        {xiao2019potential}
\bibfield{author}{\bibinfo{person}{Baicen Xiao}, \bibinfo{person}{Bhaskar
  Ramasubramanian}, \bibinfo{person}{Andrew Clark}, \bibinfo{person}{Hannaneh
  Hajishirzi}, \bibinfo{person}{Linda Bushnell}, {and} \bibinfo{person}{Radha
  Poovendran}.} \bibinfo{year}{2019}\natexlab{}.
\newblock \showarticletitle{Potential-Based Advice for Stochastic Policy
  Learning}. In \bibinfo{booktitle}{\emph{Proc. IEEE Conference on Decision and
  Control. Available: arXiv:1907.08823}}.
\newblock


\bibitem[\protect\citeauthoryear{Zhang, Torabi, Guan, Ballard, and Stone}{Zhang
  et~al\mbox{.}}{2019}]%
        {zhangleveraging}
\bibfield{author}{\bibinfo{person}{Ruohan Zhang}, \bibinfo{person}{Faraz
  Torabi}, \bibinfo{person}{Lin Guan}, \bibinfo{person}{Dana~H Ballard}, {and}
  \bibinfo{person}{Peter Stone}.} \bibinfo{year}{2019}\natexlab{}.
\newblock \showarticletitle{Leveraging Human Guidance for Deep Reinforcement
  Learning Tasks}. In \bibinfo{booktitle}{\emph{Proceedings of the
  International Joint Conference on Artificial Intelligence}}.
  \bibinfo{pages}{6339--6346}.
\newblock


\end{thebibliography}

\end{document}